\documentclass[twoside]{article}

\usepackage[accepted]{aistats2019}

\usepackage[round]{natbib}

\usepackage[utf8]{inputenc} 
\usepackage[T1]{fontenc}  
\usepackage{hyperref}       
\usepackage{url}            
\usepackage{booktabs}       
\usepackage{amsfonts}       
\usepackage{amsthm}
\usepackage{nicefrac}       
\usepackage{microtype}      
\usepackage{amsmath}
\usepackage[linesnumbered,ruled]{algorithm2e}
\usepackage{color}
\usepackage{float}
\usepackage{enumitem}
\usepackage{graphicx, xcolor}

\newtheorem{theorem}{Theorem}[section]
\newtheorem{lemma}[theorem]{Lemma}
\newtheorem{proposition}[theorem]{Proposition}

\newtheorem{definition}{Definition}[section]
\DeclareMathOperator*{\argmin}{arg\,min}
\DeclareMathOperator*{\argmax}{arg\,max}

\newcommand*\diff{\mathop{}\!\mathrm{d}}

\newcommand\independent{\protect\mathpalette{\protect\independenT}{\perp}}
\def\independenT#1#2{\mathrel{\rlap{$#1#2$}\mkern2mu{#1#2}}}

\newcommand{\Tb}{\boldsymbol{T}}
\newcommand{\Ub}{\boldsymbol{U}}

\newcommand{\Wb}{\boldsymbol{W}}
\newcommand{\Xb}{\boldsymbol{X}}

\begin{document}

\setlength{\abovedisplayskip}{3pt}
\setlength{\belowdisplayskip}{3pt}
\setlength{\abovecaptionskip}{0pt plus 0pt minus 0pt}

\runningtitle{Stabilizing Knockoffs: Multiple Simultaneous Knockoffs and Entropy Maximization}
\twocolumn[

\aistatstitle{Improving the Stability of the Knockoff Procedure: Multiple Simultaneous Knockoffs and Entropy Maximization}

\aistatsauthor{Jaime Roquero Gimenez \And James Zou }

\aistatsaddress{ Department of Statistics\\
  Stanford University\\
  Stanford, CA 94305 \\
  \texttt{roquero@stanford.edu} \\ 
  \And Department of Biomedical \\ Data Science \\
  Stanford University\\
  Stanford, CA 94305 \\
  \texttt{jamesz@stanford.edu}\\ } ]

\begin{abstract}
The \emph{Model-X knockoff} procedure has recently emerged as a powerful approach for feature selection with statistical guarantees. The advantage of knockoffs is that if we have a good model of the features $X$, then we can identify salient features without knowing anything about how the outcome $Y$ depends on $X$. An important drawback of knockoffs is its instability: running the procedure twice can result in very different selected features, potentially leading to different conclusions. Addressing this instability is critical for obtaining reproducible and robust results. Here we present a generalization of the knockoff procedure that we call simultaneous multi-knockoffs. We show that multi-knockoffs guarantee false discovery rate (FDR) control, and are substantially more stable and powerful compared to the standard (single) knockoffs. Moreover we propose a new algorithm based on entropy maximization for generating Gaussian multi-knockoffs. We validate the improved stability and power of multi-knockoffs in systematic experiments. We also illustrate how multi-knockoffs can improve the accuracy of detecting genetic mutations that are causally linked to phenotypes. 
\end{abstract}

\section{INTRODUCTION}

In many machine learning and statistics settings, we have a supervised learning problem where the outcome $Y$ depends on a subset of the features $X$, potentially in complex ways, and we would like to identify these salient features. Take medical genetics as an example, the features $X = (X_1, ..., X_d)$ are the genotypes at $d$ variants in the genome, and $Y$ is a binary indicator for the presence/absence of disease. The true model could be that $Y = f(X_\mathcal{H}, \omega)$, where $X_\mathcal{H} = \{X_i : i \in \mathcal{H}\}$ is the salient subset of the variants, and $\omega$ is some noise/randomness. It is tremendously important to identify which feature/variant is in $\mathcal{H}$.   

If we assume that $f$ is simple, say a linear function, then we might hope to use the fitted parameters of a model to select salient features. For example, we might fit a Generalized Linear Model (GLM) on $(X,Y)$ with LASSO penalty to promote sparsity in the coefficients \citep{LASSO}, and subsequently select those features with non-zero coefficients. Step-wise procedures where we sequentially modify the model is another way of doing feature selection \citep{mallowscp}. 

A clear limitation of this parametric approach is the need to to have a good model for $f$. For the genetics example, there is no great model. Moreover the standard feature selection methods are all plagued by correlations between the features: a feature that is not really relevant for the outcome, i.e. not in $\mathcal{H}$, can be selected by LASSO or Step-wise procedure, because it is correlated with relevant features. In these settings we usually lack statistical guarantees on the validity of the selected features. Finally, even procedures with statistical guarantees usually depend on having \emph{valid $p$-values}, which are based on a correct modeling of $Y|X$ and (sometimes) assume some asymptotic regime. However there are many common settings where these assumptions fail and we cannot perform inference based on those $p$-values \citep{sur2018modern}.

A powerful new approach called \emph{Model-X knockoff procedure} \citep{KN2} has recently emerged to deal with these issues. This method introduces a new paradigm: we no longer assume any model for the distribution of $Y|X$ in order to do feature selection (and therefore do not compute $p$-values), but we assume that we have full knowledge of the feature distribution $P^X$ -- or at least we can accurately model it, though there are some robustness results \citep{KN4}. This knowledge of the ground truth $P^X$ allows us to sample new \emph{knockoff} variables $\tilde{X}$ satisfying some precise distributional conditions. Although we make no assumption on $Y|X$, we can use the knockoff procedure to select features while controlling the False Discovery Rate (FDR), which is the average proportion of the selected features that are not in $\mathcal{H}$.

One major obstacle for the widespread application of knockoffs is its instability. The entire procedure sensitively depends on the knockoff sample $\tilde{X}$, which is random. Therefore, running the knockoff procedure twice may lead to very different selected sets of features. Our analysis in Section~\ref{section:experiments} shows that instability is especially severe when the number of salient features (i.e. the size of $\mathcal{H}$) is small, as is often the case. Also, whenever the number of features is very large, previous methods for generating knockoffs failed to consistently generate good samples for $\tilde{X}$, leading to inconsistent selection sets if several runs of the procedure were done simultaneously. Power also decreases drastically under those previous methods. This makes it challenging to confirm the selected variants in a replication experiment. Addressing the instability of knockoffs is therefore an important problem. 

\paragraph{Our Contributions.}
We generalize the standard (single) knockoff procedure to simultaneous multiple knockoffs (or multi-knockoffs for short). Our multi-knockoff procedure guarantees FDR control and has better statistical properties than the original knockoff, especially when the number of salient features is small. We propose a new entropy maximization algorithm to sample Gaussian multi-knockoffs. Our systematic experiments demonstrate that multi-knockoffs is more stable and more powerful than the original (single) knockoffs. Moreover we illustrate how multi-knockoffs can improve the ability to select causal variants in Genome Wide Association Studies (GWAS). 

\section{BACKGROUND ON KNOCKOFFS}\label{section:introKnockoffs}

We begin by introducing the usual setting of feature selection procedures. We consider the data as a sequence of i.i.d. samples from some unknown joint distribution: $(X_{i1},\dots,X_{id},Y_i) \sim P^{XY}$, $i = 1,\dots,N$. We then define the set of null features $\mathcal{H}_0\subset \{1,\dots,d\}$ by  $j\in \mathcal{H}_0$ if and only if $X_j \independent Y | \Xb_{-j}$ (where the $-j$ subscript indicates all variables except the $j$th and bold letters indicate vectors). The non-null features, also called alternatives, are important because they capture the truly salient effects and the goal of selection procedures is to identify them. Running the knockoff procedure gives us a selected set $\hat{\mathcal{S}}\subset \{1,\dots,d\}$, while controlling for False Discovery Rate (FDR), which stands for the expected rate of false discoveries: $FDR = \mathbb{E}\Big[ \frac{|\hat{\mathcal{S}}\cap \mathcal{H}_0|}{|\hat{\mathcal{S}}|\vee 1} \Big]$. The ratio $\frac{|\hat{\mathcal{S}}\cap \mathcal{H}_0|}{|\hat{\mathcal{S}}|\vee 1}$ is also called False Discovery Proportion (FDP).

Assuming we know the ground truth for the distribution $P^X$, the first step of the standard knockoff procedure is to obtain a \emph{knockoff} sample $\tilde{\Xb}$ that satisfies the following conditions: 

\begin{definition}[Knockoff sample]\label{def:exchange}
A knockoff sample $\tilde{\Xb}$ of a d-dimensional random variable $\Xb$ is a d-dimensional random variable such that two properties are satisfied:
\begin{itemize}[topsep=-0.25cm]
\setlength\itemsep{-0.2em}
\item Conditional independence: \hspace{10mm}$\tilde{\Xb} \independent Y | \Xb$ 
\item Exchangeability : 
\[ [\Xb,\tilde{\Xb}]_{swap(S)}\; \stackrel{d}{=}\; [\Xb,\tilde{\Xb}]\qquad  \forall S \subset \{1,\dots,d\}
\]
\end{itemize}
where the symbol $\stackrel{d}{=}$ stands for equality in distribution and $[\Xb,\tilde{\Xb}]_{swap(S)}$ refers to the vector where the original $j$th feature and the $j$th knockoff feature have been swapped whenever $j\in S$.
\end{definition}

The first condition is immediately satisfied as long as knockoffs are sampled conditionally on the sample $\Xb$ without considering any information about $Y$ (which will be the case in our sampling methods so we will not mention it again). The second condition ensures that the knockoff of each feature is sufficiently similar to the original feature in order to be a good comparison baseline. We also denote by $\Xb$ the $N\times d$ matrix where we stack the $N$ i.i.d. $d$-dimensional samples into one matrix (this is acceptable as the i.i.d. assumption allows for all sampling procedures to be done sample-wise).

The next step of the knockoff procedure constructs what we call \emph{feature statistics} \mbox{$\Wb=(W_1,\dots,W_d)$}, such that a high value for $W_j \in \mathbb{R}$ is evidence that the $j$th feature is non-null. Feature statistics described in \cite{KN2} depend only on ${[\Xb,\tilde{\Xb}] \in \mathbb{R}^{N\times 2d}},{Y\in \mathbb{R}^N}$ such that for each $j \in \{1,\dots,d\}$ we can write ${W_j = w_j([\Xb,\tilde{\Xb}],Y)} $ for some function $w_j$. The only restriction these statistics must satisfy is the \emph{flip-sign property}: swapping the $j$th feature and its corresponding knockoff feature should flip the sign of the statistic $W_j$ while leaving other feature statistics unchanged. More formally, for a subset $S\subset \{1,\dots,d\}$ of features, denoting $[\Xb,\tilde{\Xb}]_{swap(S)}$ the data matrix where the original $j$th variable and its corresponding knockoff have been swapped whenever $j\in S$, we have:
\begin{equation*}
  w_j([\Xb,\tilde{\Xb}]_{swap(S)},\!Y) \! =\!\begin{cases}
   - w_j([\Xb,\tilde{\Xb}],\!Y) , &\hspace{-1mm} \text{if $j\in S$}.\\
   w_j([\Xb,\tilde{\Xb}],\!Y) , & \hspace{-1mm} \text{otherwise}.
  \end{cases}
\end{equation*} As suggested in \cite{KN2}, the choice of feature statistics can be done in two steps: first, find a statistic 
\[
\bar{\Tb}=\bar{\Tb}([\Xb,\tilde{\Xb}],Y)=(T_1,\dots,T_d,\tilde{T}_1, \dots,\tilde{T}_d)\in \mathbb{R}^{2d}
\] 
where each coordinate corresponds to the ``importance'' --- hence we will call them \emph{importance scores} ---  of the corresponding feature (either original or knockoff). For example, $T_j$ would be the absolute value of the regression coefficient of the $j$th feature. 

After obtaining the importance score for the original and knockoff feature, we take the difference to compute the feature statistic $W_j = T_j-\tilde{T}_j$. The intuition is that importance scores of knockoffs serve as a control, a larger importance score of the $j$th feature compared to that of its knockoff implies a larger positive $W_j$ (and therefore is evidence against the null). Given some target FDR level $q \in (0,1)$ that we fix in advance, we define the selection set $\hat{\mathcal{S}}$ based on the following threshold $\hat{\tau}$:
\vspace{-2mm}
\begin{align}
\hat{\tau} =& \min \Big\{ t > 0 : \frac{1+ \#\{j: W_j \leq -t\}}{\#\{j: W_j\geq t\} \vee 1} \leq q\Big\}\label{eqn:knockoff_selection1} 
\\
\hat{\mathcal{S}} =& \{j: W_j \geq \hat{\tau}\}\label{eqn:knockoff_selection2}
\end{align}
According to Theorem 3.4 in \cite{KN2}, this procedure controls FDR at level $q$ (actually called Knockoff~+ procedure). The mechanism behind this procedure is the Selective SeqStep+ introduced in \cite{KN1}. The intuition is that we try to maximize the number of selections while bounding by $q$ an upwardly biased estimate of the FDP, which is the fraction:
\vspace{-2mm}
\begin{equation}\label{FDPestimate}
\widehat{FDP}_{KN+} = \frac{1+ \#\{j: W_j \leq -t\}}{\#\{j: W_j\geq t\}\vee 1} 
\end{equation}
The added constant in the numerator is called the ``offset'', equal to one in our case; a different FDP estimate with offset equal to 0 leads to a slightly different procedure that controls a modified, less stringent version of the FDR.
\vspace{-2mm}
\paragraph{Instability of Knockoffs} If we generate multiple replication datasets ---multiple versions of $(\Xb, Y)$, each of which is sampled from the common $P^{XY}$---then the knockoff procedure guarantees that \emph{on average}, the proportion of false discoveries is less than the desired threshold. However for a particular dataset $(\Xb, Y)$, the selected features could be very different from that of another sampled dataset, as we empirically demonstrate in Section~\ref{section:experiments}. There are settings where in half of the experiments, the knockoff procedure selects a large number of features, and it selects zero features the other half of the times. This instability is a major issue if we want to ensure that the discoveries from data are reproducible. 

The instability in the selected features is partially due to the randomness in $(\Xb, Y)$ but also in the knockoff sample $\tilde{\Xb}$. The knockoff procedure, equations~\ref{eqn:knockoff_selection1}-\ref{FDPestimate}, is sensitive to the sample $\tilde{\Xb}$. The knockoff selection set is based on a conservative estimate of the FDP given by equation (\ref{FDPestimate}). The threshold that determines the selected set requires such FDP estimate to be below some target FDR level $q$ set in advance, which in turn requires us to select at least $\lceil \frac{1}{q} \rceil$ features due to the presence of the offset in the numerator. This requirement is a great source of instability of the knockoff procedure: whenever the number of non-nulls is close to that threshold value $\lceil \frac{1}{q} \rceil$, we can end up either selecting a fairly large number of non-nulls, or not selecting any, even when the signal is strong. Our goal is to develop a new knockoff procedure which controls FDR and is more stable. We achieve this by introducing simultaneous multiple knockoffs, called multi-knockoffs for short, which extends the standard knockoff procedure.

\section{SIMULTANEOUS MULTIPLE KNOCKOFFS}

\paragraph{A Naive Flawed Approach to Multi-knockoffs} One approach to improve the stability of the selected features is to run the standard knockoff procedure multiple times in parallel and to take some type of consensus. This approach is flawed and does not control FDR. The reason is that by running knockoff multiple times in parallel, the symmetry between $\Xb$ and the knockoff samples is broken. To maintain symmetry and guarantee FDR, we need to simultaneously sample multiple knockoffs. We will make this more precise now. 

\subsection{Multi-knockoff Selection Procedure}

Fix an positive integer $\kappa \geq 2$, the multi-knockoff parameter (the usual single knockoff case corresponds to $\kappa = 1$, for which all of our results are also valid). The goal is to extend the previous distributional properties of knockoffs of Definition~\ref{def:exchange} to settings where we simultaneously sample $\kappa$ knockoff copies $(\Xb^k)_{1\leq k \leq \kappa}$ of the same $\mathbb{R}^d$-valued dataset $\Xb$ (where, again, $\Xb$ denotes either the $\mathbb{R}^d$-valued random variable when making distributional statements or a $\mathbb{R}^{N\times d}$ matrix when referring to the feature set of $N$ i.i.d. samples). As in the single knockoff setting, we can define an equivalence notion and notation for swapping multiple vectors. Instead of defining $[\Xb,\tilde{\Xb}]_{swap(S)}$ for some subset $S \subset \{1,\dots, d\}$ of indices, we consider a collection $\sigma = (\sigma_i)_{1\leq i \leq d}$ of $d$ permutations $\sigma_i$ over the set of integers $\{0,1,\dots,\kappa\}$, one for each of the $d$ initial dimensions. Whenever we will use multi-knockoffs, we will index the original features $\Xb$ by $\Xb^0$. We define the permuted vector $[\Xb^0, \Xb^1,\dots \Xb^{\kappa} ]_{swap(\sigma)} := [\Ub^0,\Ub^1,\dots,\Ub^{\kappa}]$, where $U_i^k = X_i^{\sigma_i(k)}$ for all $1\leq i \leq d$, $0\leq k \leq \kappa$. Each $\sigma_i$ permutes the $\kappa + 1$ features corresponding to the $i$th dimension of each vector $\Xb^k$, leaving the other dimensions unchanged. Once this generalized swap notion is defined, we extend the exchangeability property based on the invariance of the joint distribution to such transformations.

\begin{definition}\label{def:extendedexchange}
We say that the concatenated vectors $[\Xb^0, \Xb^1,\dots \Xb^{\kappa}]$ satisfy the extended exchangeability property if the equality in distribution $[\Xb^0, \Xb^1,\dots \Xb^{\kappa} ]_{swap(\sigma)} \stackrel{d}{=} [\Xb^0, \Xb^1,\dots \Xb^{\kappa}]$ holds for any $\sigma$ as defined above. 
\end{definition}

\begin{definition}\label{def:multiknockoff}
We say that $(\Xb^1,\dots,\Xb^{\kappa})$ is a multi-knockoff vector of $\Xb^0$ (or that they are $\kappa$ multi-knockoffs of $\Xb^0$) if the joint vector $(\Xb^0,\Xb^1,\dots,\Xb^{\kappa})$ satisfies extended exchangeability and the conditional independence requirement $(\Xb^1,\dots,\Xb^{\kappa}) \independent Y | \Xb^0$.
\end{definition}
We will later on give examples of how to generate such multi-knockoffs. We state a lemma that is a direct generalization of Lemma 3.2 in \cite{KN2} and give a proof in Appendix~\ref{appendix:proof:lemma:nullinvariance}.

\begin{lemma}\label{lemma:nullinvariance}
Consider a subset of nulls $S \subset \mathcal{H}_0$. Define a generalized swap $\sigma$ as above, where $\sigma_i$ is the identity permutation whenever $i\notin S$, and otherwise can be any permutation. Then we have the following equality in distribution for a multi-knockoff:
\[
([\Xb^0,\Xb^1,\dots,\Xb^{\kappa}],\!Y) \!\stackrel{d}{=}\! ([\Xb^0,\Xb^1,\dots,\Xb^{\kappa}]_{swap(\sigma)},\!Y)
\] 
\end{lemma}
Once the multi-knockoff vector is sampled, consider the joint vector $(\Xb^0,\Xb^1,\dots,\Xb^{\kappa})$ which takes values in  $\mathbb{R}^{(\kappa\! +\! 1)d}$. As for the single knockoff setting, we construct importance scores $\bar{\Tb} = (\Tb^0,\Tb^1,\dots,\Tb^{\kappa})$ where each $\Tb^k$ is a $d$-dimensional vector with non-negative entries. The importance scores are associated to the features in the following sense: if we generate importance scores on a swapped joint vector (as in Definition~\ref{def:extendedexchange}), then we obtain the same result as if we had swapped the importance scores of the initial joint vector. That is, the function defining the importance scores must satisfy ${[\Tb^0\!,\Tb^1\!,\dots,\!\Tb^{\kappa}]_{swap(\sigma)}\! = \!\bar{\Tb}([\Xb^0\!, \Xb^1\!,\!\dots,\! \Xb^{\kappa} ]_{swap(\sigma)},Y)}$. Common examples of such constructions are the absolute values of the coefficients associated to each feature when regressing $Y$ on $(\Xb^0,\Xb^1,\dots,\Xb^{\kappa})$, with eventually a $L^1$ penalty for sparsity. Denoting an ordered sequence by indexing in parenthesis (i.e. for any real-valued sequence $(a_0,\dots,a_n)$, we have $a_{(0)}\geq a_{(1)} \geq \dots \geq a_{(n)}$), we can define feature-wise ordered importance scores $(T_i^{(k)})_{0\leq k \leq \kappa}$ for each feature $1\leq i \leq d$. For all $1\leq i \leq d$, define: 
\begin{equation*}
\kappa_i =  \, \argmax_{0\leq k \leq \kappa} T^{k}_i \qquad \tau_i =  \, T^{(0)}_i \! - \! T^{(1)}_i 
\end{equation*}
We no longer have the possibility of generating feature statistics $W_i$ by taking an antisymmetric function of the importance scores. The extension to multi-knockoffs is done through these newly defined variables by noticing the analogy that if $\kappa = 1$ (single knockoff), then ${\tau_i = |W_i|}$ and $\kappa_i$ corresponds to the sign of $W_i$ ($\kappa_i = 0$ if and only if $W_i >0$). In the single knockoff setting, the crucial distributional result is that, conditionally on $|W_i|$, the signs of the null $W_i$ are i.i.d. flip coins. In the multi-knockoff case, the information encoded by the sign of $W_i$ is contained in $\kappa_i$, which indicates whether among a given dimension $i$ the original feature has a higher importance score than that of its knockoffs. In Appendix~\ref{appendix:intuitionKappa} we provide a geometric explanation for such choices of $\kappa_i, \tau_i$. The crucial result is that null $\kappa_i$ behave uniformly and independently in distribution and can be used to estimate the number of false discoveries. 

\begin{lemma}\label{lemma:symmetry}
The random variables $(\kappa_i)_{i\in \mathcal{H}_0}$ are i.i.d. distributed uniformly on the set $\{0,1,\dots,\kappa\}$, and independent of the remaining variables $(\kappa_i)_{i \notin \mathcal{H}_0}$, and of the feature-wise ordered importance scores $[(T_i^{(k)})_{0\leq k \leq \kappa}]_{1\leq i \leq d}$. In particular, conditionally on the variables $(\kappa_i)_{i \notin \mathcal{H}_0}$ and $(\tau_i)_{1\leq i \leq d}$, the random variables $(\kappa_i)_{i\in \mathcal{H}_0}$ are i.i.d. distributed uniformly on $\{0,1,\dots,\kappa\}$.
\end{lemma}

We prove this lemma in Appendix~\ref{proof:lemma:symmetry}. Following the steps that build the knockoff procedure as a particular case of the SeqStep+ procedure, we construct the following threshold $\hat{\tau}$ that defines the rejection set $\hat{\mathcal{S}}$ of our multi-knockoff procedure based on a FDP estimate 
\[
\widehat{FDP}_{\kappa KN+} = \frac{\frac{1}{\kappa} + \frac{1}{\kappa}\#\{ i \in \{1,\dots,d\}, \, \kappa_i \geq 1,\, \tau_i \geq t\}}{\#\{i \in \{1,\dots,d\}, \, \kappa_i = 0 ,\, \tau_i \geq t\}\vee 1}
\]
Essentially, the multi-knockoff procedure returns the features $i$ where the original feature has higher importance score than any knockoffs (i.e. $\kappa_i = 0$), and the gap with the 2nd largest importance score is above some threshold.
\vspace{-1mm}
\begin{algorithm}
    \SetKwInOut{Input}{Input}
    \SetKwInOut{Output}{Output}
    \Input{Concatenated vector $\bar{\Tb} = (\Tb^0,\Tb^1,\dots,\Tb^{\kappa})$ of importance scores, target FDR level $q$}
    \Output{Set of selected features $\hat{\mathcal{S}}$}
    \For{$i = 1$ \KwTo $d$}{$
    \kappa_i =  \, \arg\max_{0\leq k \leq \kappa} T^{k}_i \; ,
    \quad \tau_i =  \, T^{(0)}_i \! - \! T^{(1)}_i $
    }    
    {$\hat{\tau} =  \, \min \big\{t > 0 : \frac{\frac{1}{\kappa} + \frac{1}{\kappa}\#\{ i \in \{1,\dots,d\}, \, \kappa_i \geq 1,\, \tau_i \geq t\}}{\#\{i \in \{1,\dots,d\}, \, \kappa_i = 0 ,\, \tau_i \geq t\}\vee 1} \leq  q \big\}
    $}
    \Return{$\hat{\mathcal{S}} =   \{i \in \{1,\dots,d\}, \, \kappa_i = 0, \, \tau_i \geq \hat{\tau}\}$}
    \caption{Multi-knockoff Selection Procedure}\label{algorithm:MultiknockoffSelection}
\end{algorithm}
\vspace{-1mm}

\begin{proposition}\label{prop:fdrcontrol}
Fix a target FDR level $q \in (0,1)$. The procedure that selects the features in the set $\hat{\mathcal{S}}$ given by Algorithm~\ref{algorithm:MultiknockoffSelection} controls FDR at level $q$.
\end{proposition}
We prove this result in Appendix~\ref{proof:prop:fdrcontrol}. One advantage of the multi-knockoff selection procedure lies on the new value of the offset parameter. By averaging over the $\kappa$ multi-knockoffs, we are able to decrease the threshold of minimum number of rejections from $\lceil \frac{1}{q}\rceil$ to $\lceil \frac{1}{q\kappa}\rceil$, leading to an improvement in power and stability. We call $\lceil \frac{1}{q\kappa}\rceil$ the detection threshold of the multi-knockoff.  We experimentally confirm such results in Section~\ref{section:experiments}.

\subsection{Gaussian Multi-knockoffs Based on Entropy Maximization}

Most of the research and applications have focused around generating standard knockoffs when $\Xb$ comes from a multivariate Gaussian distribution $\mathcal{N}(\mu, \Sigma)$, although more universal sampling algorithms exist, for which we provide in Appendix~\ref{appendix:samplingMultiKN} a generalization to multi-knockoffs. Here we extend the existing procedures for Gaussian knockoffs to generate Gaussian multi-knockoffs for $\kappa \geq 2$. A sufficient condition for ${(\Xb^1,\dots,\Xb^{\kappa}) \in \mathbb{R}^{d\kappa}}$ to be a multi-knockoff vector is for ${(\Xb^0,\Xb^1,\dots,\Xb^{\kappa}) \in \mathbb{R}^{d(\kappa +1)}}$ to be jointly Gaussian such that: 1) all the $\Xb^k$ has the same mean $\mu$; and 2) the covariance matrix has the form :
\begin{align*}
\Sigma_\kappa =
\underbrace{
\left( 
\begin{array}{cccc}
\Sigma & \Sigma -D & \dots & \Sigma -D \\
\Sigma-D & \Sigma  & \dots & \Sigma -D \\
\vdots & \vdots  & \ddots & \vdots  \\
\Sigma-D & \Sigma -D & \dots & \Sigma \\
\end{array}
\right)
}_{\kappa +1 \; \text{blocks}}
\end{align*}
where $D$ is a diagonal matrix chosen so that $\Sigma_\kappa$ is positive semi-definite to ensure that it is a valid covariance. 

\begin{proposition}
If ${(\Xb^0,\Xb^1,\dots,\Xb^{\kappa}) \in \mathbb{R}^{d(\kappa +1)}}$ has the mean and covariance structure given above, then $(\Xb^1,\dots,\Xb^{\kappa})$ is a valid $\kappa$ multi-knockoff of $\Xb^0$.
\end{proposition}
\vspace{-1mm}
If a diagonal term $D_{ii}$ is zero, then $\Xb^k_i = \Xb^0_i$ for $k \geq 1$. This generates a valid multi-knockoff but it has no power to discover the $i$th feature (regardless of whether it is null or non-null) since each multi-knockoff is indistinguishable from the original feature. The general intuition is that the more independent the knockoffs are from the original $\Xb^0$, the greater the power of discovering the non-null features \citep{KN2}. Therefore previous work for the standard single knockoff (corresponding to $\kappa=1$) has focused on finding $D$ as large as possible in some sense, while maintaining the positive semi-definiteness of the covariance matrix. 

To construct $D$ for Gaussian multi-knockoffs, we propose maximizing the entropy $H(\Xb^0,\Xb^1,\dots,\Xb^{\kappa})$ (which has a simple closed form for Gaussian distributions). This is equivalent in the single knockoff case to minimizing mutual information, as suggested in \cite{KN2}. Indeed, $I(\Xb,\tilde{\Xb}) = H(\Xb)+ H(\tilde{\Xb}) - H(\Xb,\tilde{\Xb})$, and $H(\Xb) =  H(\tilde{\Xb})$ do not depend on $D$, hence the equivalence. 

\textbf{Entropy Knockoffs} The diagonal matrix $D(s)= diag(s_1,\dots,s_d)$ for constructing entropy multi-knockoffs is given by the following convex optimization problem:
\vspace{-3mm}
\begin{align*}
&\hspace{-7mm}\argmin_s \;  -\log \det (\frac{\kappa + 1}{\kappa}\Sigma - D(s)) - \kappa \sum_{i=1}^d \log(s_i)
\\[-1em] &\hspace{-7mm} \text{subject to}  \;
\begin{cases}
\frac{\kappa + 1}{\kappa}\Sigma - D(s) \succ 0
\\ s_i \geq 0 \qquad \forall i \in \{1,\dots,d\}
\end{cases}
\end{align*}
This optimization problem is a convex optimization problem, by noticing that $s \mapsto  -\log \det (\frac{\kappa + 1}{\kappa}\Sigma - D(s))$ is convex. It can be solved efficiently and our implementation is based on the Python package CVXOPT \citep{CVXOPT}. This knockoff construction method avoids solutions where diagonal terms are extremely close to $0$, and we provide the following lower bound on the diagonal terms of $D$: 
\begin{align*}
\big(\lambda_{min}(s)\big)^{\frac{1}{\kappa - 2}} \leq \min_{1\leq j \leq d}s_j 
\end{align*}
where $\lambda_{min}(s)$ is the smallest (positive) eigenvalue of $\frac{\kappa + 1}{\kappa}\Sigma - D(s)$. The fact that we maximize the value of the determinant of such matrix implies that we avoid having any extremely small eigenvalue, hence this bound proves useful. We provide additional analysis on the formulation of entropy maximization as a convex optimization problem and prove this lower bound in Appendix~\ref{appendix:samplingMultiKN}. Once the diagonal matrix $D$ is computed, we can generate the Gaussian multi-knockoffs by writing the conditional distribution given the original features $\Xb^0$.
\begin{align*}
& (\Xb^1, \dots, \Xb^{\kappa}) |\Xb^0 \sim    \; \mathcal{N}\big( (\boldsymbol{\mu}^1,\dots, \boldsymbol{\mu}^\kappa) , \tilde{\Sigma} \big), \;\text{where}
\\& 
\begin{cases}
\boldsymbol{\mu}^i= D\Sigma^{-1} \boldsymbol{\mu} + (I_d - D\Sigma^{-1})\Xb^0 \quad \forall 1\leq i \leq \kappa
\\ \tilde{\Sigma} = 
\left(
\arraycolsep=1.5pt\def\arraystretch{1}
\begin{array}{ccc}
C &\dots& C\!-D\! \\
C\!-D\!&\dots &C\!-D\!\\
\vdots &\vdots&\vdots \\
 C\!-D\!&\dots & C
\end{array}
\right)
\; \text{and}\; C =  2D - D\Sigma^{-1}D
\end{cases}
\end{align*}
\begin{figure}[ht]
\centering
\includegraphics[width=\linewidth]{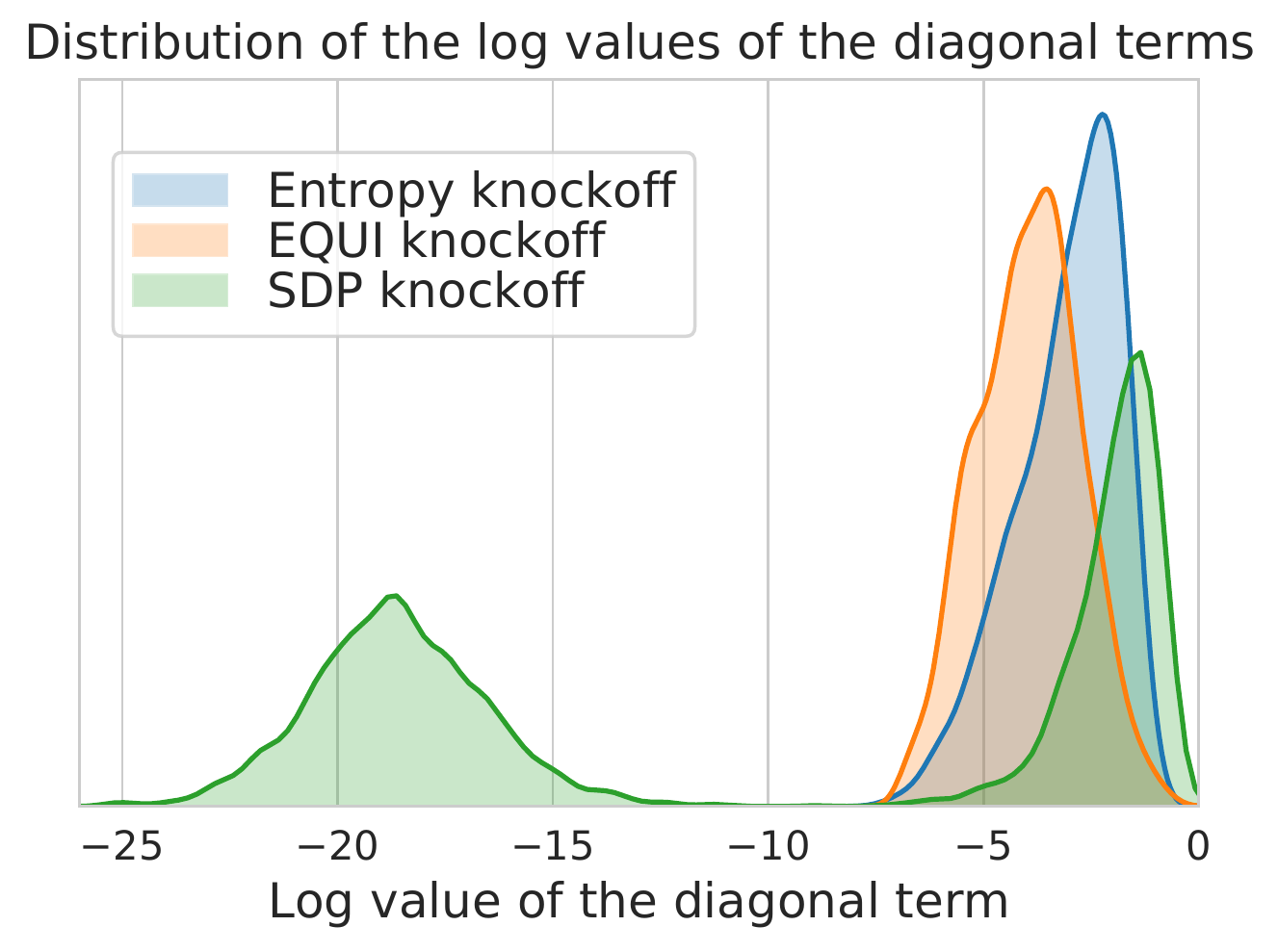}
\caption{\textbf{Comparison between methods for generating Gaussian knockoffs} We plot the densities of the distributions of the diagonal terms generated by each method. The dimension of the covariance matrix is 60.}
\label{fig:DensityDiag}
\end{figure}
In the single knockoff setting ($\kappa = 1$),  the standard approach in literature is to solve a semidefinite program (SDP) to optimize $D$. An alternative approach in the literature, called equicorrelation, is to restrict $D$ to be $D=s I_d$ and solve for $s$ (where we consider $\Sigma$ as a correlation matrix, the goal being having the same correlation between original features and knockoffs in every dimension). We provide natural generalizations of the SDP and equicorrelation to optimize the $D$ matrix for multi-knockoffs (see Appendix~\ref{appendix:samplingMultiKN}). The SDP knockoffs are based on an optimization problem that promotes sparsity: the fact that the objective function is a $L^1$-distance between the identity matrix and the diagonal matrix $D$ implies that many diagonal terms of the optimal solution will be set almost equal to 0. In addition, \cite{KN2} noticed that the equicorrelated knockoff method tends to have very low power, as in high dimensions the diagonal terms of $D$ are proportional to the lowest eigenvalue of the covariance matrix $\Sigma$, which in high-dimensional settings tends to be extremely small. Currently, SDP knockoffs are chosen by default. We perform experiments to demonstrate the advantage of entropy over SDP and equicorrelation. We randomly generate correlation matrices with the function \emph{make\_spd\_matrix} from the Python package scikit-learn \citep{scikit-learn}, and compute the diagonal matrix $D$ with the SDP and equicorrelated methods (the diagonal terms of $D$ are necessarily in the interval $(0,1)$ so that we can compare them across several runs). The lower a diagonal term is, the higher the correlation is between the original feature and its corresponding knockoff, and the less powerful is the knockoff. In Figure~\ref{fig:DensityDiag}, we plot the density of the distribution of the logarithm of the diagonal terms of $D$ (that we approximate with the empirical distribution based on 50 runs).

We see that a significant proportion of diagonal terms based on the SDP construction have values extremely small, several orders of magnitude smaller than $10^{-10}$, which effectively behave as $0$ whenever we sample knockoffs and thus the corresponding features are essentially undiscoverable because their knockoffs are too similar. In Appendix~\ref{appendix:diagonalDistribution} we show more such comparisons of the distribution of the diagonal terms for varying dimensions of the correlation matrices and strength of the correlation. More concerning in the SDP construction case, the set of almost-zero diagonal terms is very unstable to perturbations in the correlation matrix.  We report in Appendix~\ref{appendix:JaccardSimilarity} the simulations proving the instability of such sets. The outcome is simple: the Jaccard similarity between two sets of SDP undiscoverable features generated from two empirical covariance matrices obtained from two batches of i.i.d. samples from the same distribution is on average very low. That is, two parallel runs of the knockoff procedure on different datasets coming from the exact same original distribution lead to different sets of undiscoverable features.

The equicorrelated construction does not suffer such issue, although the diagonal terms tend to be smaller compared to the SDP diagonal terms that are not almost $0$. The SDP construction, due to its objective function, maximizes some diagonal terms at the expense of many others that are effectively set to 0, whereas the equicorrelated construction treats all coordinates more equally. Finally, the entropy construction achieves the best performance: the diagonal terms it constructs are generally a couple of orders of magnitude higher than the equicorrelated method, and when comparing to SDP, the entropy construction does not generate almost-zero terms, so that it does not create any catastrophic knockoff. We show in Appendix~\ref{appendix:controlleddiag} a concrete example of dataset $(\Xb,Y)$ where the SDP construction diagonal terms of the non-null features are zero so that SDP knockoffs have power 0, whereas the entropy method achieves almost full power. On top of that the whole procedure will be more stable with entropy knockoffs: there is no longer a highly variable set of undiscoverable features unrelated to the response that restricts the set of possible selections. Also, both methods have equivalent runtimes as they solve similar convex optimization problems (cf. Appendix~\ref{appendix:samplingMultiKN}).

\section{EXPERIMENTS}\label{section:experiments}

We first conduct systematic experiments on synthetic data, so that we know the ground truth. For each experiment, we evaluate both the power and the stability of multi-knockoffs and the standard knockoff.  Then we evaluate the performance of knockoff on a real set  from Genome Wide Association Studies (GWAS). 

\subsection{Analyzing Improvements with Synthetic Data}

We run simulations with synthetic data to confirm the threshold phenomenon and the improvements brought by multi-knockoffs. We randomly generate a feature matrix $\Xb$ from a random covariance matrix, fix a number of non-nulls and create a binary response Y based on a logistic response of a weighted linear combination of the non-null features. Then, we sample multi-knockoffs with $\kappa = 1$ (single knockoff), $\kappa=2$ and $\kappa = 3$ from that same $\Xb$ and run the knockoff procedure based on a logistic regression to obtain a selection set, along with values for the power and an FDP. We then repeat this whole procedure 50 times to obtain estimates of the variance and get an empirical FDR. Knockoffs are generated based on the entropy construction to show that our multi-knockoff based improvement is made on top of the entropy improvement (which is only specific for Gaussian knockoffs). The dimension of the feature vectors is 100, and the signal strength is 5. Changing the signal strength affects power but does not change the comparative behavior that we observe between single and multi-knockoffs.
\begin{figure}[ht]
\centering
\includegraphics[width=\linewidth]{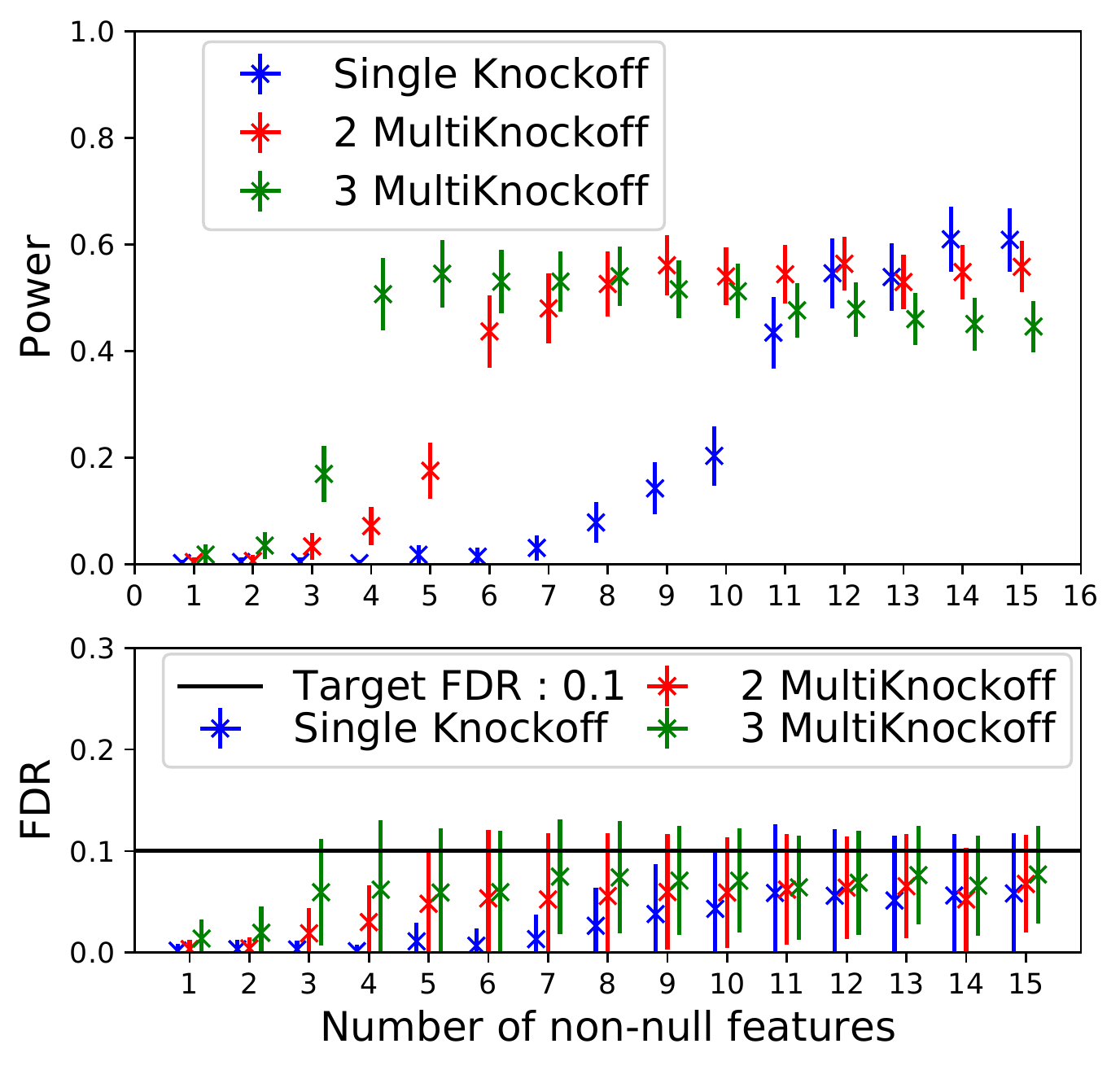}
\caption{\textbf{Power and FDR comparison between single knockoffs and multi-knockoffs} 2 and 3 multi-knockoffs has greater power than the standard knockoff when the number of non-nulls is small. All three methods control FDR. }
\label{fig:ThresholdPower}
\end{figure}

We set our target FDR level at $q = 0.1$, and compare the single knockoff setting with multi-knockoffs (with $\kappa = 2,3$) over a range of number of non-null features. We report our results in Figure~\ref{fig:ThresholdPower}. We first point out that FDR is strongly controlled in all the experiments, as expected. We then estimate the threshold values for detection given by the estimates $\lceil \frac{1}{q\kappa}\rceil$: $10$ for single knockoffs ($\kappa = 1$), $5$ rejections for multi-knockoffs with $\kappa = 2$, and $3.3$ whenever $\kappa = 3$. By plotting power as a function of the number of non-nulls we clearly confirm this threshold behavior. All three settings attain a high power regime whenever the number of non-nulls exceeds the expected detection threshold. This shows the advantage of using multi-knockoffs in settings where we expect a priori the number of non-nulls to be small, and want to make sure that our method has a chance of selecting such small set of non-nulls. We also see there is a small price to pay for using multi-knockoffs. Whenever the number of non-null features increases so that we are beyond the detection thresholds, power decreases with the number of multi-knockoffs. This is due to the fact that sampling multi-knockoffs imposes a more stringent constraint to construct the knockoff conditional distribution (cf. Appendix~\ref{appendix:samplingMultiKN}), and therefore multi-knockoffs can have slightly ``worse'' power as $\kappa$ increases.

\begin{figure}[ht]
\centering
\includegraphics[width=\linewidth]{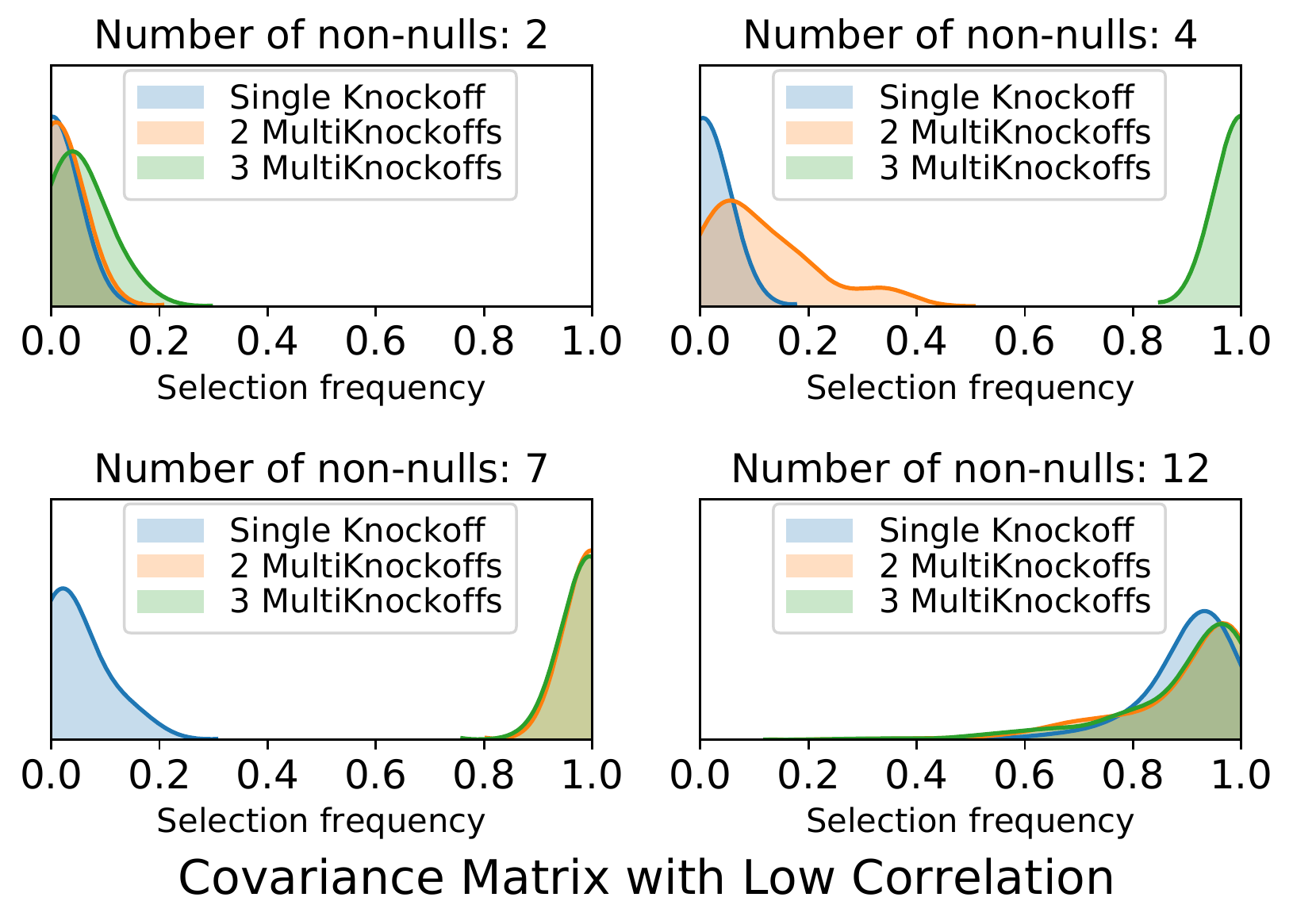}
\end{figure}
\begin{figure}[ht]
\centering
\includegraphics[width=\linewidth]{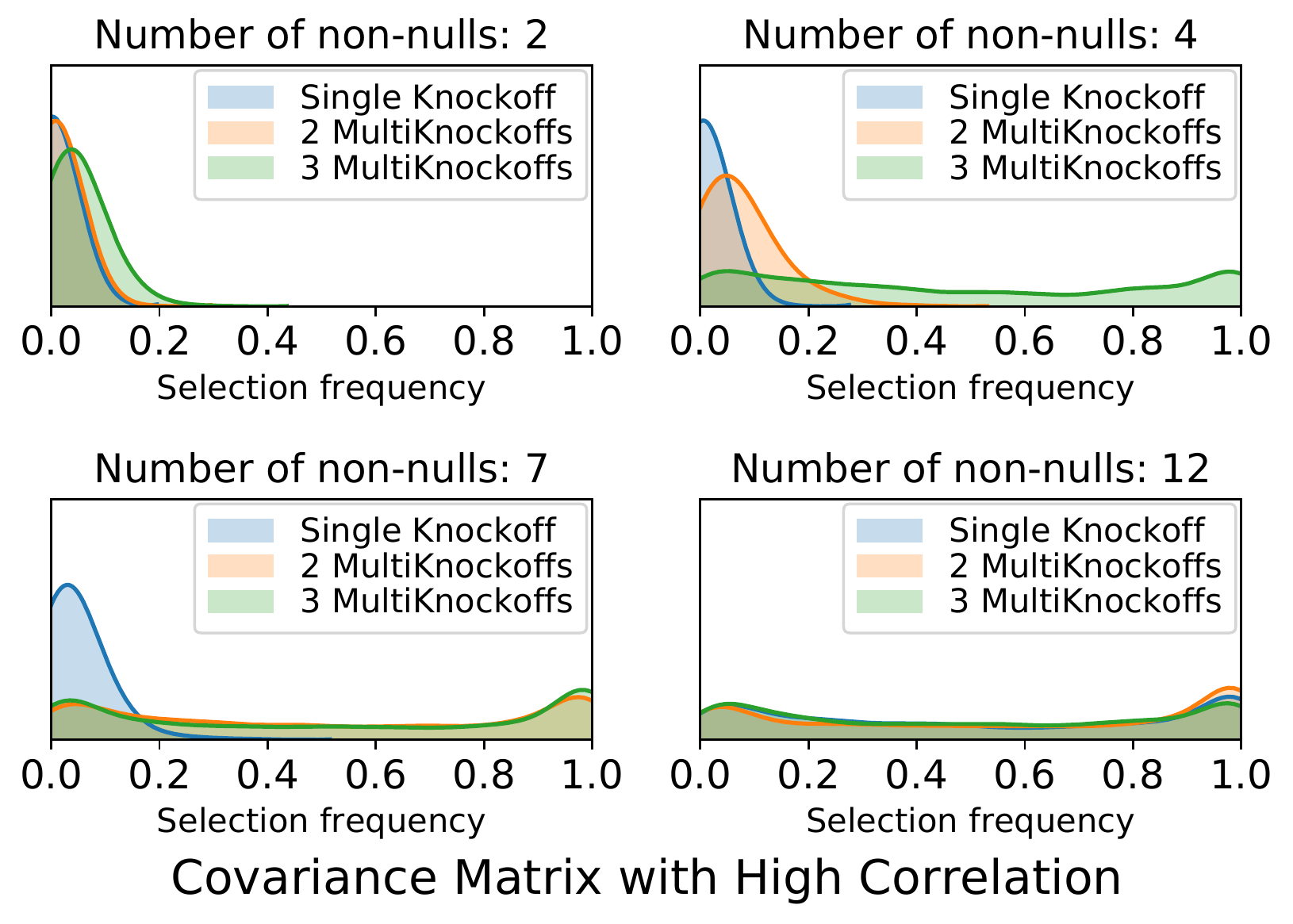}
\caption{\textbf{Improvement in stability with multi-knockoffs: density of non-nulls by selection frequency} When $X$ has low correlation setting (upper figure), and when $X$ has high correlation (lower figure).  The x-axis of each plot is the frequency that a non-null feature is selected, and the y axis indicates density.}
\label{fig:ThresholdStability}
\end{figure}

Finally, multi-knockoffs not only substantially improve the power of the procedure in settings with a small number of non-nulls; they also help stabilizing the procedure. We plot in Figure~\ref{fig:ThresholdStability}, as a function of the selection frequency, the density of the distribution of the non-nulls. In order to get the selection frequency, we run the same procedure as before, except this time we sample 200 (multi-)knockoffs out of one same $\Xb$ and run the procedure each time keeping that same $\Xb$ and the response $Y$ we had generated. This allows us to compute how frequently each non-null is selected (by repeatedly sampling knockoffs from a same $\Xb$, FDR is no longer controlled. The point of these simulations is to stress the improvement in stability). For different settings where we vary the number of non-nulls, we see that the multi-knockoffs consistently reject a large fraction of the non-nulls whenever the threshold of detection is attained. In contrast, most non-null features are selected by the standard knockoff at low frequency, indicating instability.

The key aspect here is that the improvement in power whenever a threshold is crossed is not because of an overall increase in selection frequency of all the non-nulls: the densities in the above figures do not concentrate around intermediate selection frequency values. That is, multi-knockoffs do not increase the power by increasing instability.

\subsection{Applications: GWAS Causal Variants}

We apply our stabilizing procedures for fine mapping the causal variants in a genome wide association study (GWAS). These studies scan the whole genome in search of single nucleotide polymorphisms (SNPs) that are associated with a particular phenotype. In practice, they compute correlation scores for each SNP with respect to the phenotype, and select those beyond a certain significance threshold. Often times, the high correlation between SNPs (called linkage disequilibrium) implies that a large number of consecutive SNPs have a large association score and thus are selected. Fine mapping consists in finding the precise causal SNPs that really help explain the phenotype. Knockoffs can be useful in this setting, but the threshold phenomenon described earlier is an impediment to the application of knockoffs. We want to analyze several dozens, maybe hundreds of SNPs that have passed the selection threshold of the GWAS. However, the number of true causal SNPs may be very low, possibly less than 10. If we set a target FDR level of $0.1$, the single knockoff procedure may be unable to make any detection.  

\begin{figure}[ht]
\centering
\includegraphics[width=\linewidth]{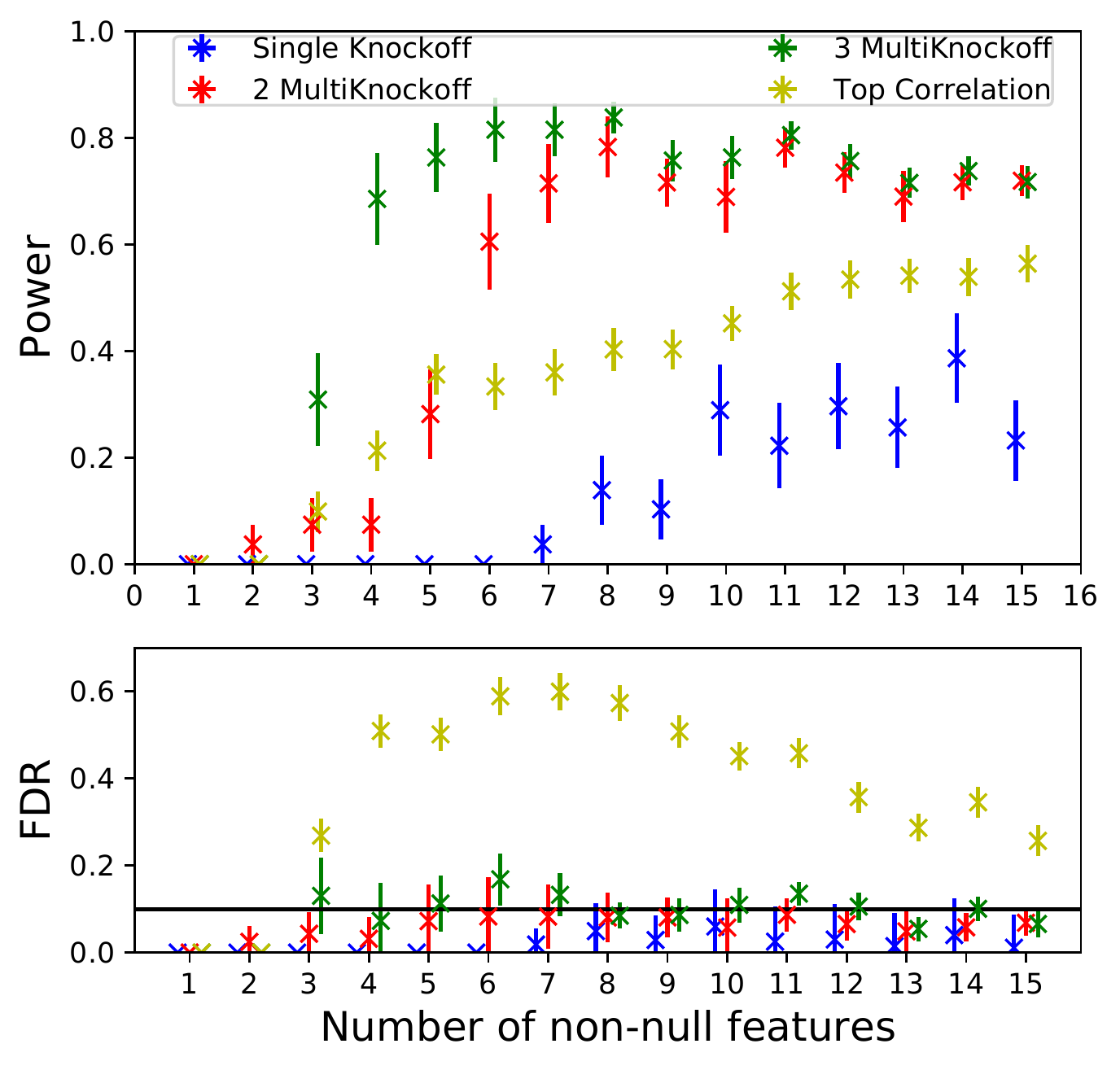}
\caption{\textbf{Power and FDR comparison between single knockoffs, multi-knockoffs, and top correlation for a GWAS dataset.} }
\label{fig:realdata}
\end{figure}

We follow the lines of \cite{eCAVIAR, CAVIAR} and run simulations analogous to those presented in Figure~\ref{fig:ThresholdPower}, where the features now correspond to individual genotypes. As it is not possible to actually know, for a given phenotype, which are the true causal SNPs without experimental confirmation, we generate synthetic responses (phenotypes) by randomly choosing a given number of SNPs as causal. Such semi-synthetic data (real $X$ and simulated $Y$) is standard in literature \citep{CAVIAR}. In addition, we run a selection procedure without statistical guarantees that is commonly used: we pick the top correlated SNPs with the response. We give more details in Appendix~\ref{appendix:detailsGWAS}, and detail the impact of the approximation assumptions on the observed FDR. We recover the results obtained with synthetic data and report them in Figure~\ref{fig:realdata}: FDR is controlled with the multi-knockoff procedure, and the top correlation method fails to control FDR. We also observe the detection threshold effect: for a low number of causal SNPs, single knockoffs have almost no power, and multi-knockoffs have better power than picking the most correlated SNPs. 

\section{DISCUSSION}
In this paper, we propose multi-knockoffs, an extension of the standard knockoff procedure. We show that multi-knockoff guarantees FDR control, and demonstrate how to generate Gaussian multi-knockoffs via a new entropy-maximization algorithm. Our extensive experiments show that multi-knockoffs are more stable and more powerful compared to the standard (single) knockoff. Finally we illustrate on the important problem of identifying causal GWAS mutations that multi-knockoff substantially outperforms the popular approach of selecting mutations with the highest correlation with the phenotype. The main contribution of this paper is in proposing the mathematical framework of multi-knockoffs; additional empirical analysis and applications is an important direction of future work. 

\subsubsection*{Acknowledgments}

J.R.G. was supported by a Stanford Graduate Fellowship. J.Z. is supported by a Chan–Zuckerberg Biohub Investigator grant and National Science Foundation (NSF) Grant CRII 1657155. The authors thank Emmanuel Candès and Nikolaos Ignatiadis for helpful discussions.

\bibliographystyle{unsrtnat}

\bibliography{main}

\newpage

\appendix

\section*{APPENDIX}

\section{SAMPLING MULTIPLE KNOCKOFFS}\label{appendix:samplingMultiKN}

\subsection{Gaussian Multi-knockoffs}

We generalize the knockoff generation procedure to have $\kappa \geq 2$ multi-knockoffs, starting with the Gaussian case. We see that a sufficient condition for ${(\Xb^1,\dots,\Xb^{\kappa}) \in \mathbb{R}^{d\kappa}}$ to be a multi-knockoff vector -besides all vectors $\Xb^{j}$ having the same mean $\mu$- is that the joint vector ${(\Xb^0,\Xb^1,\dots,\Xb^{\kappa}) \in \mathbb{R}^{d(\kappa +1)}}$ has a covariance matrix of the form:
\begin{align*}
\Sigma_\kappa =
\underbrace{
\left( 
\begin{array}{cccc}
\Sigma & \Sigma -D & \dots & \Sigma -D \\
\Sigma-D & \Sigma  & \dots & \Sigma -D \\
\vdots & \vdots  & \ddots & \vdots  \\
\Sigma-D & \Sigma -D & \dots & \Sigma \\
\end{array}
\right)
}_{\kappa +1 \; \text{blocks}}
\end{align*}
We can easily generalize previous diagonal matrix constructions to the multi-knockoff setting. The mathematical formulation of the heuristic behind SDP and equicorrelated knockoffs -as an objective function in the convex optimization problem- does not change when sampling multi-knockoffs, as the correlation between an original feature and any of its multi-knockoffs is the same as a consequence of exchangeability. However, the positive semi-definite constraint that defines the feasible set changes with $\kappa$. For the entropy knockoffs, the objective function depends also on $\kappa$.

Because all three methods solve a similar convex optimization problem, there is no significant difference in runtime.

\begin{proposition}\label{proposition:diagonalMultiKN}
We generalize the diagonal construction methods SDP, equicorrelated and entropy when sampling $\kappa \geq 2$ multi-knockoffs from a multivariate Gaussian, by the following convex optimization problems. We recover the formulations for the single knockoff setting by replacing $\kappa = 1$.
\begin{itemize}
\setlength\itemsep{-1mm}
\item \textbf{SDP Multi-knockoffs} For a covariance matrix $\Sigma$ whose diagonal entries are equal to one, the diagonal matrix $D(s)= diag(s_1,\dots,s_d)$ for constructing SDP knockoffs is given by the following convex optimization problem:
\begin{align*}
&\text{minimize} \quad  \sum_{i=1}^d |1-s_i|
\\&\text{subject to}  \;
\begin{cases}
\setlength\itemsep{-2mm}
\frac{\kappa + 1}{\kappa}\Sigma - D(s) \succ 0
\\ s_i \geq 0 \qquad \forall i \in \{1,\dots,d\}
\end{cases}
\end{align*}
\item \textbf{Equicorrelated Multi-knockoffs} For a covariance matrix $\Sigma$ whose diagonal entries are equal to one, the diagonal matrix $D(s) = s I_d$ for constructing equicorrelated knockoffs is given by the following convex optimization problem:
\begin{align*}
&\text{maximize} \quad  s \qquad \text{subject to}  \;
\begin{cases}
\frac{\kappa + 1}{\kappa}\Sigma - s I_d \succ 0
\\ s \geq 0 
\end{cases}
\end{align*}
The solution of this optimization problem has a closed form expression: $s^* = \frac{\kappa + 1}{\kappa}\lambda_{min}(\Sigma)$, where $\lambda_{min}(\Sigma)$ is the smallest (positive) eigenvalue of $\Sigma$.
\item \textbf{Entropy Multi-knockoffs} The diagonal matrix $D(s)= diag(s_1,\dots,s_d)$ for constructing entropy knockoffs is given by the following convex optimization problem (as $s \mapsto -\log \det (2\Sigma - D(s))$ is convex):
\begin{align*}
&\hspace{-7mm}\argmin_s \;  -\log \det (\frac{\kappa + 1}{\kappa}\Sigma - D(s)) - \kappa \sum_{i=1}^d \log(s_i)
\\ &\hspace{-7mm} \text{subject to}  \;
\begin{cases}
\frac{\kappa + 1}{\kappa}\Sigma - D(s) \succ 0
\\ s_i \geq 0 \qquad \forall i \in \{1,\dots,d\}
\end{cases}
\end{align*}
\end{itemize}

The entropy knockoff construction method avoids solutions where diagonal terms are extremely close to $0$, and we provide the following lower bound on the diagonal terms of $D$:
\begin{align*}
(\lambda_{min}(s))^{\frac{1}{\kappa - 2}} \leq \min_{j\in \{1,\dots, d\}}s_j 
\end{align*}
where $\lambda_{min}(s)$ is the smallest (positive) eigenvalue of $\frac{\kappa + 1}{\kappa}\Sigma - D(s)$. 
\end{proposition}

For the SDP method and the equicorrelated method, increasing the number of multi-knockoffs constrains the feasible set of the convex optimization problem. However, diagonal terms can always be as close to 0 as they want, and we empirically observe a slight decrease in power as we increase $\kappa$ indicating that the added constraints limit the choice of ``good'' values for the diagonal terms.

\begin{proof}
The heuristic behind the different construction methods looks for different optimal solutions to convex optimization problems. Depending on the multi-knockoff parameter $\kappa$, we need to adapt two parts of the convex optimization formulations: the objective function and the feasible set. Objective functions in the SDP and equicorrelated constructions remain unchanged as they do not depend on the number of multi-knockoffs.
\setlength\itemsep{-2mm}
\item
\paragraph{Adapting the Feasible Set}

We first look at how the constraints defining the feasible set change as we go from simple knockoffs to multi-knockoffs. All three methods (SDP, equicorrelated, entropy) define the feasible set for $s=(s_1,\dots, s_d)\in \mathbb{R}_{+}^d$ by constraining $\Sigma_{\kappa}$ to be positive definite. We show that this constraint is equivalent to $\frac{1 + \kappa}{\kappa}\Sigma - D \succ 0 $, which we prove by induction. Suppose that at step $\kappa \geq 1$, for any positive definite matrix $S$, for $D$ positive definite diagonal matrix,
\begin{align*}
\underbrace{
\left(
\begin{matrix}
S & S -D & \dots & S -D \\
S-D & S  & \dots & S -D \\
\vdots & \vdots  & \ddots & \vdots  \\
S-D & S -D & \dots & S \\
\end{matrix}
\right)
}_{\kappa +1 \; \text{blocks}} \succ  \:  0
\\ \Leftrightarrow \qquad \frac{1 + \kappa}{\kappa}S - D \succ 0
\end{align*}
where we write $A \succ 0 $ for $A$ symmetric positive definite. We repeatedly use the characterization of a symmetric positive definite matrix via its Schur complement. We have : 
\begin{align*}
&\Sigma_{\kappa + 1} =  
\underbrace{
\left(
\begin{matrix}
\Sigma & \Sigma -D & \dots & \Sigma -D \\
\Sigma-D & \Sigma  & \dots & \Sigma -D \\
\vdots & \vdots  & \ddots & \vdots  \\
\Sigma-D & \Sigma -D & \dots & \Sigma \\
\end{matrix}
\right)
}_{\kappa+2 \; \text{blocks}} \succ  \:  0
\end{align*}
\begin{align*}
\Leftrightarrow &
\underbrace{
\left(
\begin{array}{ccc}
\Sigma & \dots & \Sigma -D \\
\vdots & \ddots & \vdots \\
\Sigma - D & \dots & \Sigma
\end{array}
\right)
}_{\kappa +1\; \text{blocks}} 
\\ & - 
\left(
\begin{array}{c}
\Sigma - D
\\
\vdots
\\ \Sigma - D 
\end{array}
\right) \Sigma^{-1}
\left(
\begin{array}{ccc}
\!\Sigma\! - \!D \!&\dots&\! \Sigma\! -\! D
\end{array}
\right)
\succ  \: 0 
\\ \Leftrightarrow &
\underbrace{
\left(
\begin{array}{ccc}
C & \dots & C -D \\
\vdots & \ddots & \vdots \\
C - D & \dots &C
\end{array}
\right)
}_{\kappa +1\; \text{blocks}} \!
\succ   0, \; \text{$C$ defined below}
\\ \Leftrightarrow & \; 
\frac{1 + \kappa}{\kappa}C - D \succ 0,  \quad \text{by induction as $C\succ 0$}
\\ \Leftrightarrow & \;
\frac{2 + \kappa}{1+\kappa}D - D\Sigma^{-1}D \succ 0
\\ \Leftrightarrow & \;
\left(
\begin{array}{cc}
\Sigma & D
\\ D & \frac{2 + \kappa}{1+\kappa}D
\end{array}
\right) \succ 0 
\\ \Leftrightarrow & \;
\Sigma - D (\frac{2 + \kappa}{1+\kappa}D)^{-1}D \succ 0 
\\ & \text{as} \; \Sigma \succ 0 \; \text{and this is a Schur complement}
\\ \Leftrightarrow & \;
 \frac{2 + \kappa}{1+\kappa}\Sigma - D \succ 0
\end{align*}
Hence the recursive step and we conclude the proof. We have
\[{C = \Sigma - (\Sigma - D)\Sigma^{-1}(\Sigma - D)} = 2D - D\Sigma^{-1}D \succ 0
\]
given that $\Sigma \succ0$ so $C$ is the Schur complement of :
\[
\left(
\begin{array}{cc}
\Sigma & \Sigma - D
\\ \Sigma - D & \Sigma
\end{array}
\right) \succ 0
\]
\setlength\itemsep{-2mm}
\item
\paragraph{Objective Function for Entropy Construction}
In addition to this, we need to formulate the objective function for the entropy construction. The entropy of a multivariate Gaussian has a simple closed formula.
\begin{align*}
H(\Xb^0,\Xb^1,\dots,\Xb^{\kappa})= \frac{1}{2}\log \det (2\pi e \Sigma_{\kappa}) 
\end{align*}
We rearrange the expression of $\det(\Sigma_{\kappa})$ to show that minimizing $-\log(\det(2\pi e \Sigma_{\kappa}))$ is equivalent to minimizing
\[
-\log \det (\frac{\kappa + 1}{\kappa}\Sigma - D(s)) - \kappa \sum_{i=1}^d \log(s_i)
\]
(We showed in the main text that minimizing the entropy in a Gaussian setting is equivalent to minimizing this log-determinant). In order to do so, it suffices to show by induction that the following holds for all $\kappa \geq 1$:
\[
\det(\Sigma_{\kappa})  \propto \det(D)^{\kappa}\det(\frac{\kappa + 1}{\kappa}\Sigma - D)
\]
where the multiplicative constant is a real number depending only on $\kappa$. We first show this for $\kappa = 1$. 
\begin{align*}
\det(\Sigma_1) = & \det 
\left(
\begin{array}{cc}
\Sigma & \Sigma - D(s) \\
\Sigma - D(s) & \Sigma
\end{array}
\right)
\\ = & \det(\Sigma) \det\big(\Sigma - (\Sigma - D(s))\Sigma^{-1}(\Sigma - D(s))\big)
\\ = & \det(\Sigma) \det\big(2D(s) - D(s)\Sigma^{-1}D(s)\big)
\\ = & \det 
\left(
\begin{array}{cc}
\Sigma & D(s) \\
 D(s) & 2D(s)
\end{array}
\right)
\\ = & \det(2\Sigma D(s) - D(s)D(s))
\\ = & \det(2\Sigma - D(s))\prod_{i=1}^d s_i
\end{align*}

Suppose the result holds for a given $\kappa \geq 1$. We use the notation $|A| = \det(A)$. We have:
\begin{align*}
|\Sigma_{\kappa + 1}| =  &
\underbrace{
\left|
\begin{matrix}
\Sigma & \Sigma -D & \dots & \Sigma -D \\
\Sigma-D & \Sigma  & \dots & \Sigma -D \\
\vdots & \vdots  & \ddots & \vdots  \\
\Sigma-D & \Sigma -D & \dots & \Sigma \\
\end{matrix}
\right|
}_{\kappa+2 \; \text{blocks}} 
\\ = & |\Sigma|
\underbrace{
\left(
\begin{array}{ccc}
\Sigma & \dots & \Sigma -D \\
\vdots & \ddots & \vdots \\
\Sigma - D & \dots & \Sigma
\end{array}
\right)
}_{\kappa +1\; \text{blocks}} 
\\ & - 
\left(
\begin{array}{c}
\Sigma - D
\\
\vdots
\\ \Sigma - D 
\end{array}
\right) \Sigma^{-1}
\left(
\begin{array}{ccc}
\!\Sigma\! - \!D \!&\dots&\! \Sigma\! -\! D
\end{array}
\right)
\Bigg|
\end{align*}
\begin{align*}
 =& |\Sigma|
\underbrace{
\left|
\begin{array}{ccc}
C & \dots & C -D \\
\vdots & \ddots & \vdots \\
C - D & \dots &C
\end{array}
\right|
}_{\kappa +1\; \text{blocks}} 
\\ \propto & |\Sigma| |D|^{\kappa} \Big|\frac{1 + \kappa}{\kappa}C - D \Big|  \qquad \text{by induction} 
\\ \propto & |\Sigma| |D|^{\kappa}  \Big|\frac{2 + 2\kappa}{\kappa}D  - \frac{1 + \kappa}{\kappa}D\Sigma^{-1}D - D\Big|
\\ \propto & |\Sigma| |D|^{\kappa + 1} \Big|\frac{2 + \kappa}{\kappa}I - \frac{1 + \kappa}{\kappa}\Sigma^{-1}D\Big|
\\ \propto & |D|^{\kappa + 1} \Big|\frac{2 + \kappa}{\kappa+1}\Sigma - D\Big|
\end{align*}
Hence the result, where $C$ is the same as before. We used the following two formulae to compute determinants of block matrices: 
\begin{itemize}
\setlength\itemsep{-0.2em}
\item If $A$ is invertible, then 
\[\det 
\left(
\begin{array}{cc}
A & B \\
 C & D
\end{array}
\right) = \det(A)\det(D - CA^{-1}B)
\]
\item If $C$ and $D$ commute and all the blocks are square matrices, then $\det 
\left(
\begin{array}{cc}
A & B \\
 C & D
\end{array}
\right) = \det(AD -BC)$
\end{itemize}

\item
\paragraph{Lower Bound for Diagonal Terms in Entropy Construction}
For the entropy construction, in order to give a lower bound for the $s_i$, we derive an expression for the solution of the minimization problem. Without loss of generality, fix $j \in \{1,\dots,d\}$ so that we compute the partial derivative with respect to $s_j$. Denote $R(s) = \frac{\kappa + 1}{\kappa} \Sigma - D(s)$. Using Jacobi's formula for the derivative of a determinant, we get: 
\begin{align*}
 &\frac{\diff}{\diff s_j} \big( |R(s)|\prod_{i=1}^d s_i^{\kappa} \big) 
 \\ & =   \Big(\frac{\diff}{\diff s_j} |R(s)| \Big) \prod_{i=1}^d s_i^{\kappa} +  |R(s)|\big(\prod_{i\neq j}^d s_i\big) s_j^{\kappa - 1} 
\\ & =  |R(s)| tr\Big( R(s)^{-1}\frac{\diff R(s)}{\diff s_j} \Big)\! \prod_{i=1}^d s_i\! +\!  |R(s)|s_j^{\kappa - 1}\prod_{i\neq j}^d s_i
\\ & =   (s_j^{\kappa - 1}  - s_j R(s)_{jj}^{-1}) \Big( |R(s)|\prod_{i\neq j}^d s_i \Big)
\end{align*}
given that $\frac{\diff}{\diff s_j}  R(s) = - \frac{\diff}{\diff s_j} D(s) = - \mathbb{I}_{jj}$ where $\mathbb{I}_{kl}$ is a matrix where the only non-zero term equal to one is in position $(kl)$. Therefore $tr\Big( R(s)^{-1}\frac{\diff R(s)}{\diff s_j} \Big) = -R(s)^{-1}_{jj}$. Setting this expression to $0$ we get that the solution of the convex optimization problem satisfies 
\[
\frac{1}{s_j^{\kappa - 2}} = R(s)_{jj}^{-1} \qquad \forall j \in \{1,\dots,d\}
\]
Now we can write the diagonal term in the inverse matrix as a quotient between two determinants: ${R(s)^{-1}_{jj} = \frac{M_j(s)}{\det(R(s))}}$ where $M_j(s)$ is the principal minor of $R(s)$ when removing the $j$th row and column. As both $M_{j}(s)$ and $\det R(s)$ can be written as a product of eigenvalues, the Cauchy interlacing theorem gives the following lower bound: 
\[
\lambda_{min}(R(s)) \leq \min_{j\in \{1,\dots, d\}}s_j^{\kappa - 2} 
\]
where $\lambda_{min}(R(s))$ is the smallest (positive) eigenvalue of $R(s)$.
\end{proof}

\subsection{General Multi-knockoff Sampling Based on SCIP}

We can also generalize to the multi-knockoff setting a universal (although possibly intractable) knockoff sampling algorithm introduced in \cite{KN2}: the Sequential Conditional Independent Pairs (SCIP). Fix $\kappa \geq 1$ the number of multi-knockoffs to sample (so that SCIP corresponds to $\kappa = 1$). We iterate for $1\leq i \leq d$ over the features, at each step sampling $\kappa$ knockoffs for the $i$th feature, independently one of another, from the conditional distribution of the original feature given all the available variables sampled so far. It is important to notice that, whenever SCIP is tractable due to the particular structure of a given initial feature distribution (as for Hidden Markov Models), this generalization to multi-knockoffs will also be tractable given that increasing the number of multi-knockoffs does not alter the conditional dependencies between knockoffs and original features. We formulate this in Algorithm~\ref{algorithm:SCIPmulti} and prove that the resulting samples satisfy exchangeability. 

\begin{algorithm}
    \For{$1\leq i\leq d$}{
    \For{$1\leq k \leq \kappa$}{
    Sample $X^k_i \sim \mathcal{L}(X^0_i | \Xb_{-i}^0, X_{1:i-1}^{1:\kappa})$
    \\ }
    }
    \Return{$X^{1:\kappa}_{1:d}$}
    \caption{Sequential Conditional Independent Multi-knockoffs}\label{algorithm:SCIPmulti}
\end{algorithm}
\begin{proof}
We need to prove the following equality in distribution, using the notations of Definition~\ref{def:extendedexchange}: 
\begin{equation*}
[\Xb^0, \Xb^1,\dots \Xb^{\kappa} ]_{swap(\sigma)} \stackrel{d}{=} [\Xb^0, \Xb^1,\dots \Xb^{\kappa}]
\end{equation*}
We follow the same proof as in \cite{KN2}, where we have the following induction hypothesis:
\paragraph{Induction Hypothesis:} After i steps, we have
\[
[\Xb^0, \Xb^1_{1:i},\dots \Xb^{\kappa}_{1:i} ]_{swap(\sigma)} \stackrel{d}{=} [\Xb^0, \Xb^1_{1:i},\dots \Xb^{\kappa}_{1:i}]
\]
where now $\sigma = (\sigma_j)_{1\leq j \leq i}$ with arbitrary permutations $\sigma_j$ over $\{0,\dots,\kappa\}$. After the first step the equality holds for $i=1$ given that all $X^k_1$ have the same conditional distribution and are independent one of another. Now, if the hypothesis holds at step $i-1$, then at step $i$ we have that the joint distribution of $[\Xb^0, \Xb^1_{1:i},\dots \Xb^{\kappa}_{1:i}]$ can be decomposed as a product of conditional distributions given the sampling procedure so that we have: 
\begin{align*}
    \mathcal{L}(\Xb^0, \Xb^1_{1:i},\dots \Xb^{\kappa}_{1:i}) = \frac{ \prod_{k=0}^{\kappa} \mathcal{L}(X^k_i | \Xb_{-i}^0, X_{1:i-1}^{1:\kappa})}{\mathcal{L}( \Xb_{-i}^0, X_{1:i-1}^{1:\kappa})}
\end{align*}
Now, by induction hypothesis, the expression in the denominator satisfies the extended exchangeability for the $i-1$ first dimensions (we marginalize out over $X^0_i$ which doesn't matter as at step $i-1$ the permutations $\sigma_j$ are over $j\leq i-1$). And so are the terms in the numerator, as again we permute only elements among the first $i-1$ dimensions. And, because of the conditional independent sampling, the numerator expression is also exchangeable for the $i$th dimension. In conclusion, $\mathcal{L}(\Xb^0, \Xb^1_{1:i},\dots \Xb^{\kappa}_{1:i})$ is exchangeable for the first $i$ dimensions, hence concluding the proof.
\end{proof}

\section{PROOFS}

\subsection{Proof of Lemma~\ref{lemma:nullinvariance}}\label{appendix:proof:lemma:nullinvariance}

\begin{proof}
Given that the $swap(\sigma)$ operation is the concatenation of the action of each permutation $\sigma_i$ onto $(X_i^0,\dots, X_i^{\kappa})$ and that we can write $\sigma_i$ as the composition of transpositions, we see that it is enough to show the result for a simple transposition of two features (original or multi-knockoff) corresponding to a null dimension. This leads us directly to the proof of Lemma 3.2 in \cite{KN2}, where the difference is that we add all the extra multi-knockoffs in the conditioning set.
\end{proof}

\subsection{Proof of Lemma~\ref{lemma:symmetry}}
\begin{proof}\label{proof:lemma:symmetry}
Consider any collection $(\sigma_i)_{i\in \mathcal{H}_0}$ of permutations $\sigma_i$ on the set $\{0,\dots, \kappa\}$, and for $i\notin \mathcal{H}_0$, set $\sigma_i = ()$ the identity permutation. In order to prove the result we need to show the following equality in distribution:
\begin{align*}
\Big( [\sigma_i(\kappa_i)]_{1\leq i \leq d},&[(T_i^{(k)})_{0\leq k \leq \kappa}]_{1\leq i \leq d} \Big)
\\& \stackrel{d}{=} \Big([\kappa_i]_{1\leq i \leq d},[(T_i^{(k)})_{0\leq k \leq \kappa}]_{1\leq i \leq d} \Big)
\end{align*}
Define $\hat{T}_i^{k} = T_i^{\sigma_i(k)}$ for every $i \in \{1,\dots, d\}$ and $k\in \{0,\dots, \kappa\}$. Using the notation for the extended swap this is equivalent to $\hat{\Tb} = \Tb_{swap(\sigma)}$, where for each null index $i\in \mathcal{H}_0$ the $i$th features of $T$ and its knockoffs have been permuted according to $\sigma_i$ (and the non-null remained at their place). By construction, $\Tb = \Tb(\Xb,Y)$ is a function of $\Xb$ and $Y$ which associates to each feature in $\Xb$ a ``score'' for its importance (for simplicity here we will denote by $\Xb$ the whole concatenated vector of $[\Xb^0, \Xb^1,\dots \Xb^{\kappa}]$). The choice of such function is restricted so that $\Tb_{swap(\sigma)} = \Tb(\Xb_{swap(\sigma)},Y)$. By the multi-knockoff exchangeability property, and our specific choice of $\sigma$ that does not permute non-null features, we also have $(\Xb_{swap(\sigma)},Y) \stackrel{d}{=} (\Xb, Y)$. This in turn implies:
\[
\hat{\Tb} \stackrel{d}{=} \Tb
\]
Also, given that the permutation is done feature-wise, the feature-wise ordered importance scores remain the same. 
\[
[(\hat{T}_i^{(k)})_{0\leq k \leq \kappa}]_{1\leq i \leq d} = [(T_i^{(k)})_{0\leq k \leq \kappa}]_{1\leq i \leq d}
\]
We now prove the equality in distribution (where we have an abusive notation for representing set probabilities):
\begin{align*}
    & \mathbb{P}([\kappa_i]_{1\leq i \leq d},[(T_i^{(k)})_{0\leq k \leq \kappa}]_{1\leq i \leq d}) 
    \\ &=  \, \mathbb{P}([ T_{i}^{\kappa_i}\! =\! T_{i}^{(0)}]_{1\leq i \leq d},[(T_i^{(k)})_{0\leq k \leq \kappa}]_{1\leq i \leq d} )
    \\ &=  \, \mathbb{P}([\hat{T}_{i}^{\kappa_i} \!=\! \hat{T}_{i}^{(0)}]_{1 \leq i \leq d},[(\hat{T}_i^{(k)})_{0\leq k \leq \kappa}]_{1\leq i \leq d} )
    \\ &=  \, \mathbb{P}( [T_{i}^{\sigma_i(\kappa_i)} \!=\! T_{i}^{(0)}]_{1\leq i \leq d},[(T_i^{(k)})_{0\leq k \leq \kappa}]_{1\leq i \leq d} )
    \\ &=  \, \mathbb{P}([\sigma_i(\kappa_i)]_{1\leq i \leq d},\; [(T_i^{(k)})_{0\leq k \leq \kappa}]_{1\leq i \leq d} )
\end{align*}

The second equality is due to the equality in distribution between $\Tb$ and $\hat{\Tb}$, and the third equality makes use of the fact that for any $i \in \{1,\dots,d\}$ the order statistics of $(\hat{T}_i^{0},\dots,\hat{T}_i^{\kappa})$ and $(T_i^{0},\dots,T_i^{\kappa})$ are the same. The statement about our variables $\tau_i$ holds because they are functions of the feature-wise ordered importance scores. 
\end{proof}

\subsection{Proof of Proposition~\ref{prop:fdrcontrol}}
\begin{proof}\label{proof:prop:fdrcontrol}
The random variables $\kappa_i$ allow us to construct one-bit p-values as in \cite{KN1}, while the $\tau_i$ can be used to determine the ordering in which we sort those p-values, given that conditionally on $(\tau_i)_{1\leq i \leq d}$, we have $(\kappa_i)_{i\in \mathcal{H}_0}$ i.i.d. uniform over $\{0,\dots,\kappa\}$, independent of $(\kappa_i)_{i\notin\mathcal{H}_0}$. We can therefore permute the dimension indices based on $(\tau_i)_i$ so that $\tau_1 \geq \tau_2 \geq \dots \geq \tau_d \geq 0$, and still define the following random variables with the desired properties. We expect that our ordering based on $(\tau_i)_{i}$ will tend to place non-nulls at the beginning. Set for $1\leq i \leq d$:
\[
p_i = 
\begin{cases}
\frac{1}{\kappa + 1},  &\kappa_i = 0
\\ 1,  &\kappa_i \geq 1
\end{cases}
\]
The distributional results for $(\kappa_i)_{i\in \mathcal{H}_0}$ imply that the null $(p_i)_{i\in \mathcal{H}_0}$ are also i.i.d., independent of the non-null $(p_i)_{i\notin \mathcal{H}_0}$ and the $(\tau_i)_{1\leq i \leq d}$ and have the following distribution:
\[
\begin{cases}
\mathbb{P}(p_i = \frac{1}{\kappa +1}) = \frac{1}{\kappa + 1}
\\ \mathbb{P}(p_i = 1) = \frac{\kappa}{\kappa + 1}
\end{cases}
\]
In particular, null $p_i$ satisfy $p_i \stackrel{d}{\geq} \mathcal{U}([0,1])$. Fix a target FDR level $q \in (0,1)$, and a constant $c\in (0,1)$. Following \cite{KN1}, define the Selective SeqStep+ threshold:
\[
\hat{k} = \max \Bigg\{1\leq k \leq d, \;  \frac{1+\#\{i \leq k : p_i > c\}}{\#\{i\leq k : p_i \leq c\} \vee 1} \!\leq\! \frac{1\!-\!c}{c}q \Bigg\}
\]
Then according to Theorem 3 in \cite{KN1}, the procedure that selects the features ${\mathcal{S} = \{i \leq \hat{k},\, p_i \leq c\}}$, controls for FDR at level $q$. For the particular choice of $c = \frac{1}{\kappa +1}$, we have:
\begin{align*}
\hat{k} = & \max \Bigg\{\!1\leq k \leq d,   \frac{1+\#\{i \leq k : p_i > \frac{1}{\kappa +1}\}}{\#\{i\leq k : p_i \leq \frac{1}{\kappa +1}\} \vee 1} \!\leq\! \kappa q \Bigg\}
\\ = & \max \Bigg\{\!1\leq k \leq d,   \frac{ 1 + \# \{i \leq k : \kappa_i \geq 1\}}{\#\{i\leq k : \kappa_i = 0\} \vee 1} \leq \kappa q \Bigg\}
\\ = & \max \Bigg\{\!1\leq k \leq d,   \frac{ \frac{1}{\kappa} \!+\! \frac{1}{\kappa}\# \{i : \kappa_i \geq 1,\, \tau_i \geq \tau_k \}}{\#\{i : \kappa_i = 0,\, \tau_i \geq \tau_k \} \vee 1}\!\leq\!  q \Bigg\}
\end{align*}

Now, instead of maximizing over $k$ indexing a decreasing sequence $\tau_1\geq \dots \geq \tau_d$, one can formulate the problem as minimizing the threshold $\tau$:
\[
\tau^* \!=\!  \min \!\Bigg\{\!\tau > 0,\! \frac{\frac{1}{\kappa}\! +\! \frac{1}{\kappa}\# \{1\!\leq \!i\! \leq \!d: \kappa_i \!\geq \!1,\! \tau_i \!\geq \!\tau \}}{\#\{1\!\leq \!i\!\leq \!d: \kappa_i\! = \!0,\! \tau_i \!\geq \!\tau \} \vee 1} \!\leq\! q \Bigg\}
\]
The selection set is then defined as:
\[
\hat{\mathcal{S}} =   \{i \in \{1,\dots,d\}, \, \kappa_i = 0, \, \tau_i \geq \hat{\tau}\}
\]
\end{proof}

We notice that the main role of $\tau_i$ is to determine an ordering sequence of the p-values for the Adaptive SeqStep+ procedure. Any function of the ordered statistics $(T_i^{(k)})_{0\leq k \leq \kappa}$ gives valid statistics that can be used to order the p-values, given that the distributional restrictions will still be satisfied. A rich literature covers this topic \citep{lei2018adapt, lei2017star, ignatiadis2016data}, and could be applied to multi-knockoff based p-values.

\subsection{Intuition for Choice of Kappa and Tau}\label{appendix:intuitionKappa}

We illustrate the particular choice of $(\kappa_i)$ and $(\tau_i)$ from a geometric point of view. For the single knockoffs, one can pair the importance statistics of each original feature and its knockoff $(T_i,\tilde{T}_i)$ and plot such pairs as points in a plane $\mathbb{R}_{+}^2$. We then have a geometric view of the threshold selection. Consider the parallel lines given by the equations $y= x + t$ and $y = x - t$, partitioning the plane into 3 sections. The terms $\#\{j: W_j \leq -t\}$ and $\#\{j: W_j\geq t\}$ in the FDP estimate

\begin{equation*}
\widehat{FDP}_{KN+} = \frac{1+ \#\{j: W_j \leq -t\}}{\#\{j: W_j\geq t\}\vee 1} 
\end{equation*}

are obtained by counting the number of points $(T_i,\tilde{T}_i)$ in the section above $y= x + t$ (that is, $y \geq x + t$) and below $y = x - t$ (that is, $y \leq x - t$). For $t=0$, the two lines collapse and $\mathbb{R}^2_{+}$ is partitioned by the line $y=x$.

The same setting can happen in higher dimensions, where we partition the space $\mathbb{R}^d$ into $d$ cones given by $C_i = \{x \in \mathbb{R}_{+}^d, x_i = \max_j x_j\}$. Our method for choosing a threshold for multi-knockoffs proceeds as before: for a given $t>0$, we count the number of points $(T^{0}_i,T^1_i,\dots,T^{\kappa}_i)\in \mathbb{R}^{\kappa +1}$ in each translated cone $C_{it} = \{x \in \mathbb{R}_{+}^d, x_i \geq t +  \max_{j\neq i} x_j\}$ and compare the counts in $C_{0t}$ corresponding to the original feature to the average over those in $C_{it}$. We then find the minimum $t$ subject to some constraint. Reformulating this gives our variables $\kappa_i$ and $\tau_i$.

\section{SUPPLEMENT ON SIMULATIONS}

\subsection{Comparison Between Distributions of Diagonal Construction Methods}\label{appendix:diagonalDistribution}

\begin{figure}[ht]
\centering
\includegraphics[width=\linewidth]{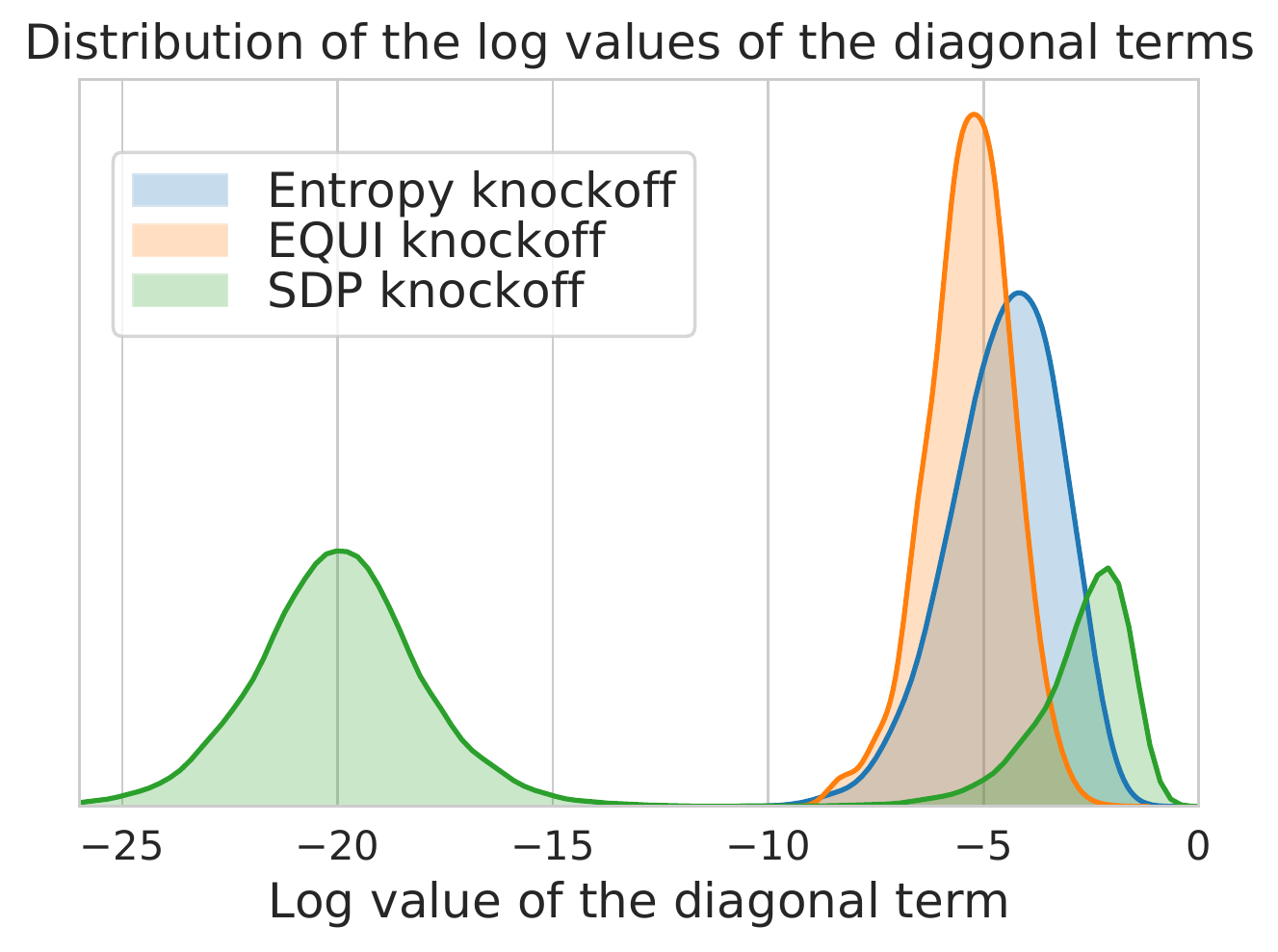}
\caption{\textbf{Comparison Between Diagonal Matrix Construction Methods - Increased Dimension to 400} }
\label{fig:diagDistrHighDim}
\end{figure}

We run another simulation where we increase the dimension of the samples. We plot again the distribution of the logarithm of the diagonal terms for the three construction methods in Figure~\ref{fig:diagDistrHighDim}. As we increase the dimension, we observe that the distributions are shifted towards more negative values, indicating that the diagonal coefficients constructed tend to be smaller. This is particularly the case for the equicorrelated construction. The SDP construction generates an even higher proportion of almost-zero diagonal terms as we increase the dimension. Also, increasing the level of correlation has also an impact on the distribution of the diagonal terms similar to what we observe by increasing the dimension.

\subsection{Measuring Stability of the Set of SDP-based Undiscoverable Features with Jaccard Similarity}\label{appendix:JaccardSimilarity}

\begin{figure}[ht]
\centering
\includegraphics[width=\linewidth]{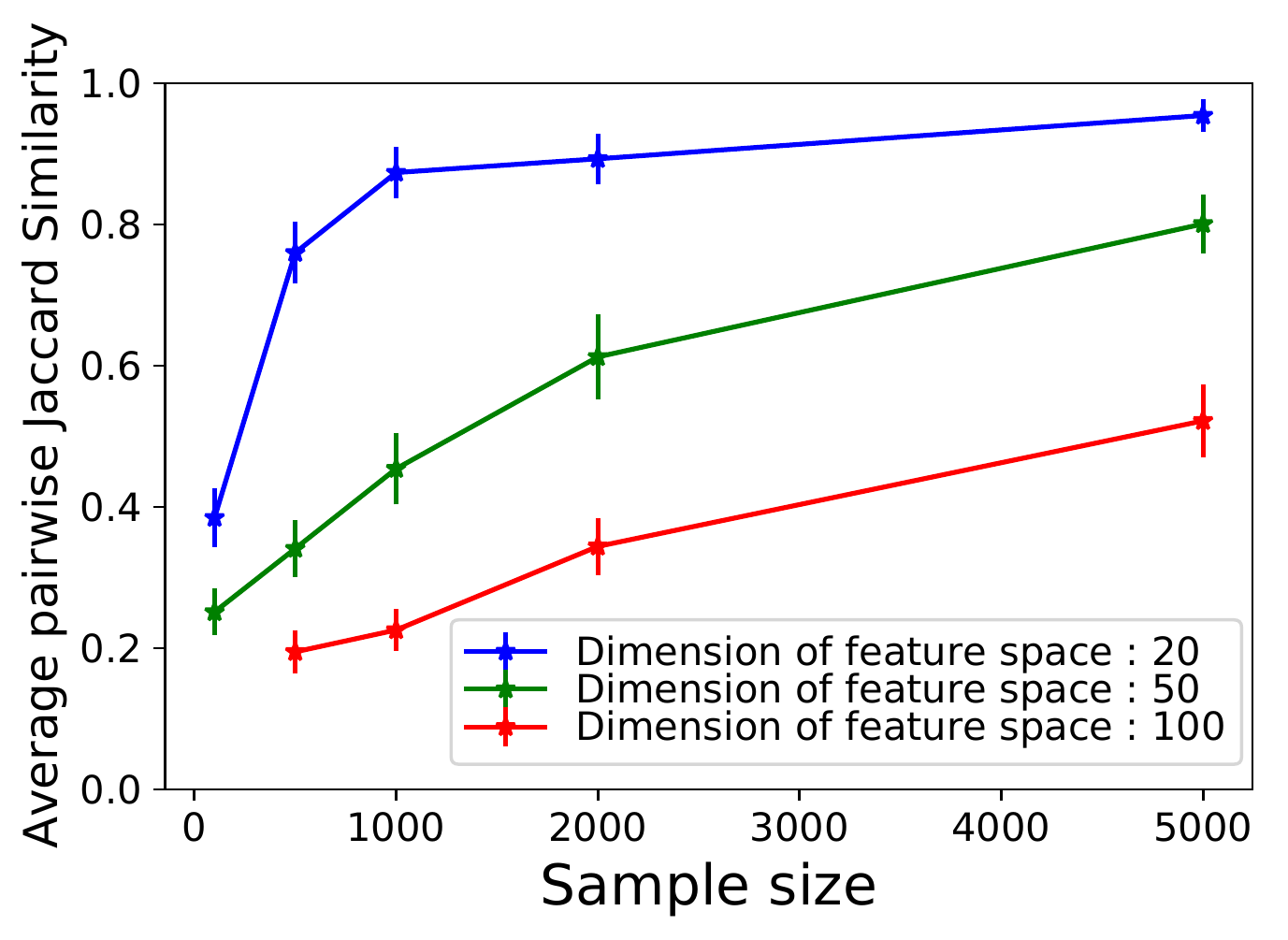}
\caption{\textbf{Average Pairwise Jaccard Similarity for Multiple Runs of SDP Method} }
\label{fig:jaccard}
\end{figure}

For a given correlation matrix, we generate samples from a centered multivariate Gaussian. Based on the estimated correlation matrix from these samples, we run the SDP construction to get the matrix $D$, and identify the set of undiscoverable features. By repeatedly doing this, we obtain multiple sets of undiscoverable features. In Figure~\ref{fig:jaccard} we plot the averaged Jaccard similarity over all pairs of such sets, as a function of the sample size (and repeat the whole procedure 50 times to estimate the variance of our results). Even though the similarity increases with the sample size, it remains very low. Furthermore, the similarity decreases with the dimension, so in high-dimensional problems where $d >> N$ then the SDP construction method is very unstable, and has a very high proportion of undiscoverable features as suggested in Figure~\ref{fig:diagDistrHighDim}. Reproducing findings becomes then very hard in such settings if we use SDP knockoffs.

\subsection{Comparing Power Between SDP and Entropy Knockoffs}\label{appendix:controlleddiag}

We show an extreme example of the drastic improvement in power brought by entropy knockoffs over SDP knockoffs. We generate a dataset $(\Xb,Y)$ where we specify the distribution of the feature set such that we can predict which diagonal coefficients will be set to 0 in the SDP method, and thus construct the response $Y$ such that the non-null features are undiscoverable. We choose a particular covariance structure that conveniently allows for explicit expressions of the diagonal terms in each method, though the results apply more generally as shown in Figure~\ref{fig:DensityDiag}.

We sample $\Xb \sim \mathcal{N}(0, \Sigma)$ as a multivariate centered Gaussian random variable, where $\Sigma \in \mathbb{R}^{3d\times 3d}$ is a covariance matrix defined as a block-diagonal matrix:
\[
\Sigma = \underbrace{
\left(
\begin{array}{cccccc}
A & 0 & 0 & \dots & 0 & 0 \\
0 & A & 0 & \dots & 0 & 0 \\
0 & 0 & A & \dots & 0 & 0 \\
\vdots & \vdots & \vdots & \ddots & \vdots & \vdots \\
0 & 0 & 0 & \dots & A & 0 \\
0 & 0 & 0 & \dots & 0 & A
\end{array}
\right)
}_{d \; \text{blocks}} 
\]
where $A = \left(
\begin{array}{ccc}
1 & a & 0  \\
a & 1 & a  \\
0 & a & 1  
\end{array}
\right)$ for some $a\geq 0$. 

\begin{figure}[ht]
\centering
\includegraphics[width=\linewidth]{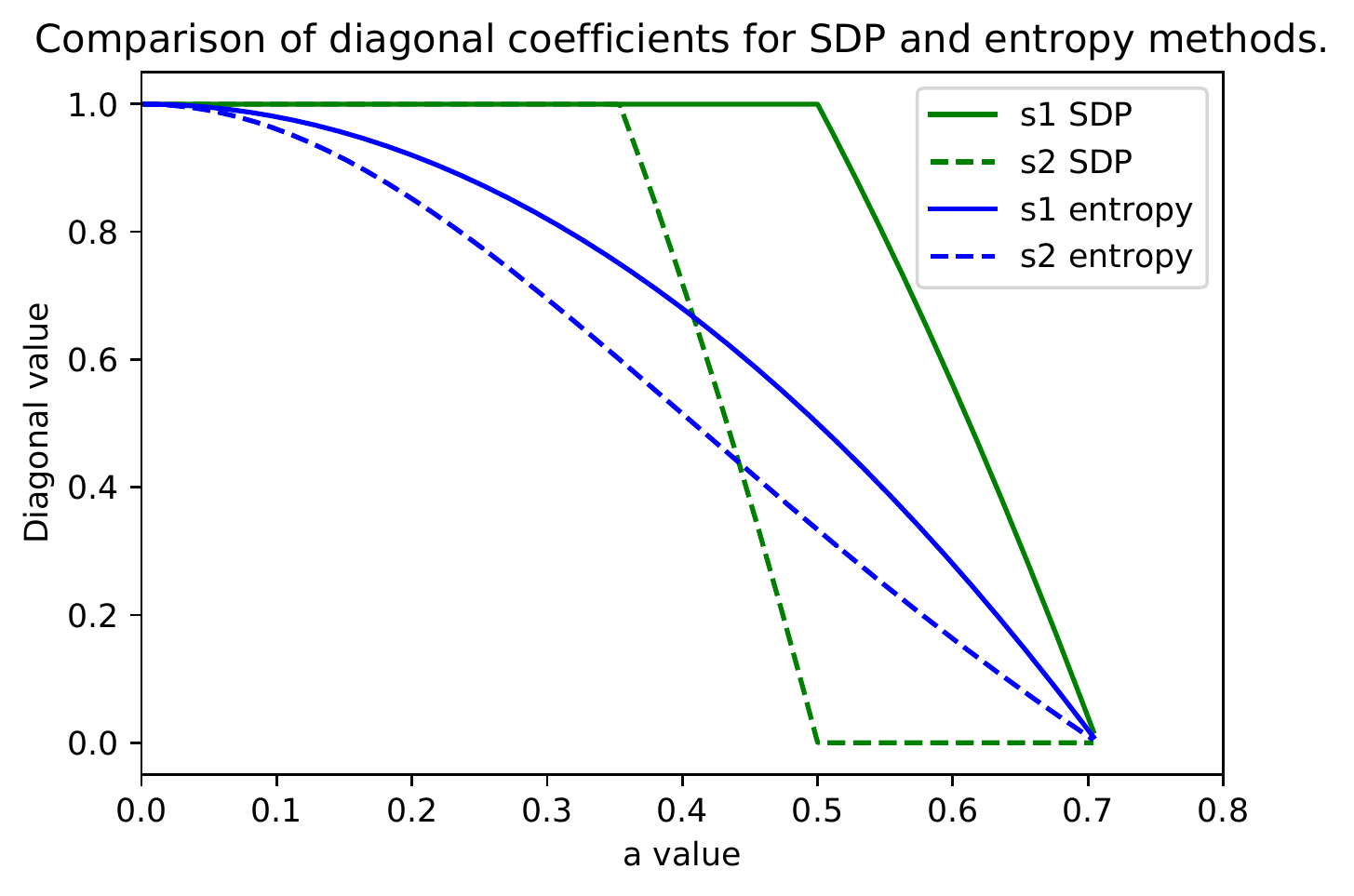}
\caption{\textbf{Comparison of Diagonal Values for SDP and Entropy Methods} For values of $a$ in the range $[\frac{1}{2},\frac{1}{\sqrt{2}})$ the value $s_2^{SDP}(a)$ is exactly equal to 0: the optimization objective favors setting $s_2^{SDP}$ to 0 in order to maximize $s_1^{SDP}$. Entropy knockoffs do not suffer from this issue.}
\label{fig:diagonalcoeff}
\end{figure}

SDP and entropy methods output a diagonal matrix $D = sI_{3d}$ such that $s \in \mathbb{R}^{3d}$ is the concatenation $d$ times the sequence $(s_1,s_2,s_1)\in \mathbb{R}^3$, which corresponds to the output of the corresponding method on the matrix $A$. We can derive an explicit formula for $s_1, s_2$ as functions of $a$ for both the SDP and entropy methods, which we denote $(s_1^{SDP}(a), s_2^{SDP}(a))$ and $(s_1^{entr}(a), s_2^{entr}(a))$. We plot such curves in Figures~\ref{fig:diagonalcoeff}, which show that for a wide range of values of $a$, $s_2^{SDP}(a)$ is exactly equal to 0, whereas the diagonal terms of the entropy method stay always positive. Notice that in this particular setting the maximal value that $a$ can take is $\frac{1}{\sqrt{2}}$, otherwise the convex optimization problem has an empty feasible set.

This phenomenon becomes worse as we increase the number of simultaneously correlated features, we refer again to Figures~\ref{fig:DensityDiag} and \ref{fig:diagDistrHighDim}.

We now generate a large number of samples so that the estimated empirical correlation matrix is very close to the real one (so that sample size is not a factor when comparing SDP and entropy methods). We then sample a response vector $Y$ such that the non-null features correspond to the dimensions associated to the $s_2$ diagonal terms (i.e. the non-null features are given by $\mathcal{H}_0= \{3i + 2, \; 0\leq i \leq (d-1)\}$).
The non-null features are therefore undiscoverable under the SDP construction, whereas entropy knockoffs are still able to select the non-nulls. The results of simulating the whole procedure are clear: SDP has zero power, and entropy knockoffs have full power. Of course, this is an extreme situation designed to accentuate this behavior. Still, across the multiple simulations done in this paper, entropy knockoffs consistently had higher power than SDP knockoffs.

\subsection{Generating the Synthetic Response for the Real Genome Dataset}\label{appendix:detailsGWAS}

We collect data from the 1000 Genomes Project \citep{1000genomes}, and obtain around 2000 individual samples for 27 distinct segments of chromosome 19 containing an average of 50 SNPs per segment. We filter out SNPs that are extremely correlated (above 0.95), and generate for each of those 27 segments a random subset that will correspond to the causal SNPs. We then generate the response accordingly and use a logistic regression to obtain importance scores. For the top correlation method, we select the top $k$ correlated features, where $k$ is chosen as the number of rejections that multi-knockoffs make, so that we have a fair comparison between methods.

One important caveat that explains why sometimes the averaged FDP is above the target is that with real data, it is crucial to accurately estimate the feature distribution. In these simulations, we approximate the 0/1/2 matrix of SNPs by a Gaussian distribution, where we need to estimate the covariance based on the data. Such inaccurate approximation causes the average FDP to exceed the target sometimes. However, the knockoff procedure is robust to mis-estimations of the feature distribution \citep{KN4}, so that we can expect FDR control at an inflated level. Our FDR results are therefore satisfactory, and the comparison is stark with the top correlation method that catastrophically fails to control FDR. 

\end{document}